\documentclass[letterpaper]{article} 
\usepackage{aaai2026}  
\usepackage{times}  
\usepackage{helvet}  
\usepackage{courier}  
\usepackage[hyphens]{url}  
\usepackage{graphicx} 
\usepackage{natbib}  
\usepackage{caption} 
\usepackage{algorithm}
\usepackage{algorithmic}

\usepackage{newfloat}
\usepackage{listings}
\DeclareCaptionStyle{ruled}{labelfont=normalfont,labelsep=colon,strut=off} 
\floatstyle{ruled}
\newfloat{listing}{tb}{lst}{}
\floatname{listing}{Listing}
\usepackage{array}  

\usepackage{arydshln} 
\usepackage{pifont} 
\usepackage{amssymb} 
\usepackage{multirow} 
\usepackage{booktabs} 
\usepackage{longtable} 
\usepackage{CJKutf8} 
\usepackage{enumitem}
\usepackage{fixltx2e}
\usepackage{amsmath}

\usepackage{xcolor} 
\definecolor{customgreen}{HTML}{82B366} 
\definecolor{customorange}{HTML}{D79B00} 
\definecolor{customred}{HTML}{B85450} 
\definecolor{customblue}{HTML}{6C8EBF} 
\definecolor{customgreen}{HTML}{82B366} 

\title{Investigating the Impact of Rationales for LLMs on Natural Language Understanding}
\author{
    Wenhang Shi\textsuperscript{\rm 1}, 
    Shuqing Bian\textsuperscript{\rm 2},
    Yiren Chen\textsuperscript{\rm 3},
    Xinyi Zhang\textsuperscript{\rm 1},
    Zhe Zhao\textsuperscript{\rm 2}
    Pengfei Hu\textsuperscript{\rm 2},
    Wei Lu\textsuperscript{\rm 1},
    Xiaoyong Du\textsuperscript{\rm 1}\thanks{Corresponding Author}
}
\affiliations{
    \textsuperscript{\rm 1} Renmin University of China
    
    \textsuperscript{\rm 2}Tencent
    
    \textsuperscript{\rm 3}Peking University\\

    \{wenhangshi, xinyizhang.info, lu-wei, duyong\}@ruc.edu.cn, yrchen92@pku.edu.cn, shuqingbian@gmail.com, \\ \{nlpzhezhao, alanpfhu\}@tencent.com

}

\begin{document}

\nocopyright
\maketitle

\begin{abstract}
Chain-of-thought (CoT) rationales, which provide step-by-step reasoning to derive final answers, benefit LLMs in both inference and training. 
Incorporating rationales—either by generating them before answering during inference, or by placing them before or after the original answers during training—significantly improves model performance on mathematical, symbolic and commonsense reasoning tasks. 
However, most work focuses on the role of rationales in these reasoning tasks, overlooking their potential impact on other important tasks like natural language understanding (NLU) tasks.
In this work, we raise the question: \textbf{Can rationales similarly benefit NLU tasks?} 
To conduct a systematic exploration, we construct NLURC, a comprehensive and high-quality NLU dataset collection with rationales, and develop various rationale-augmented methods.
Through exploring the applicability of these methods on NLU tasks using the dataset, we uncover several potentially surprising findings:
(1) CoT inference shifts from hindering NLU performance to surpassing direct label prediction as model size grows, indicating a positive correlation.
(2) Most rationale-augmented training methods perform worse than label-only training, with one specially designed method consistently achieving improvements.
(3) LLMs trained with rationales achieve significant performance gains on unseen NLU tasks, rivaling models ten times their size, while delivering interpretability on par with commercial LLMs.

\end{abstract}

\section{Introduction}
Chain-of-thought (CoT) prompting, which guides large language models (LLMs) to generate step-by-step reasoning during inference, has been shown to significantly improve model performance on complex reasoning tasks such as mathematical, symbolic and commonsense reasoning \cite{DBLP:conf/nips/KojimaGRMI22,DBLP:conf/nips/Wei0SBIXCLZ22}.
Recent works combine this capability with model distillation, eliciting CoT rationales from large teacher models and incorporating them into student model training as additional supervision signals \cite{DBLP:journals/corr/HintonVD15, DBLP:conf/acl/HoSY23}.
The most common method places rationales before or after the answer, resulting in substantial improvements on reasoning tasks compared to answer-only training \cite{DBLP:conf/acl/LiHYRC023,DBLP:conf/icml/FuPOSK23,DBLP:conf/emnlp/WadhwaAW24}.
Additionally, CoT Collection augments existing instruction-tuning datasets with rationales, further enhancing model generalization performance on unseen tasks, particularly challenging ones \cite{DBLP:conf/emnlp/KimJKJYSS23}.

While the role of rationales in inference and training for mathematical or symbolic reasoning tasks has been well studied, their impact on other important tasks like natural language understanding (NLU) remains underexplored. NLU tasks, such as sentiment analysis and paraphrase detection, require models to locate and analyze key information from input to produce structured output \cite{DBLP:journals/corr/abs-2006-03541,DBLP:journals/csur/MinaeeKCNCG21}. Unlike reasoning tasks that rely on step-by-step decomposition to derive answers, NLU tasks can often be solved by understanding  key parts of input. Due to their ease of control, automation and scalability, NLU tasks are widely applied in real-world scenarios, making them important evaluation dimension for LLMs
 \cite{DBLP:journals/tmlr/LiangBLTSYZNWKN23,DBLP:conf/emnlp/QinZ0CYY23}.

In this work, we raise the question: \textbf{Whether incorporating rationales into the inference and training of NLU tasks bring improvements, and if so, which methods are most effective.} One might intuitively assume that the same rationale-based approaches that enhance reasoning tasks would also boost NLU performance. But we find that NLU tasks do not benefit from rationales in the same way.


To systematically investigate the impact of rationales on NLU tasks, we construct both rationale-augmented NLU datasets and methods. 
Specifically, we introduce the NLU Rationale Collection (NLURC), a comprehensive and high-quality NLU dataset collection with 252,507 rationales, covering 34 widely used datasets across 11 NLU task types. 
In addition, we develop a range of rationale-augmented training and inference methods, including those that have been empirically validated as effective in  reasoning tasks.
By comparing and analyzing the applicability of these methods in NLU scenarios using NLURC, we uncover several, sometimes counter-intuitive, findings regarding impact of rationales on inference, training, and unseen tasks of NLU.
We summarize our key findings as follows:
\vspace{-0.08cm}
\begin{itemize}[itemsep=2pt, parsep=0pt]
\item \textbf{As model size grows, the impact of CoT inference shifts from hindering to enhancing performance.}
While NLU tasks can often be solved by focusing on key parts of the input, we find that CoT reasoning in NLU tends to analyze the entire text before providing the answer. 
For smaller models with limited capacity, CoT's over-analysis increases the chance of understanding errors, which actually harms performance.
\item \textbf{Most methods of incorporating rationales into training degrade performance compared to label-only training.} Common practice of placing rationales before or after answers offer no benefit. Only the method of computing separate losses for labels and rationales consistently outperforms, highlighting the crucial role of label learning in NLU tasks.
\item \textbf{Incorporating rationales into label-only instruction-tuning data further enhances LLM generalization.} It improves performance on unseen NLU tasks, rivaling models ten times larger, and boosts interpretability by improving the quality of model-generated rationales, comparable to commercial LLMs.
\end{itemize}



\section{Related Work}
\subsection{Inference with Rationales}
Chain-of-thought (CoT) prompting encourages LLMs to generate intermediate reasoning steps before arriving at a final answer, significantly enhancing performance on complex reasoning tasks \cite{DBLP:conf/nips/KojimaGRMI22, DBLP:conf/nips/Wei0SBIXCLZ22}.
Various studies propose different variants of CoT prompting, such as organizing reasoning within hierarchical, programmatic, and graph-based frameworks \cite{DBLP:conf/nips/YaoYZS00N23, DBLP:journals/tmlr/ChenM0C23, DBLP:conf/aaai/BestaBKGPGGLNNH24}, or explicitly generating multiple reasoning paths, combined with strategies like multi-agent debate or majority votes \cite{DBLP:conf/icml/Du00TM24, DBLP:conf/iclr/0002WSLCNCZ23}.
These test-time scaling methods further increase computational costs and improve performance on reasoning tasks \cite{DBLP:journals/corr/abs-2408-03314}. 
In this work, we explore whether generating reasoning first boosts NLU performance, leaving more advanced CoT methods for future work.

\subsection{Training with Rationales}
Small models often struggle to generate reasoning chains and  leverage inference with rationales to improve performance \cite{DBLP:conf/nips/Wei0SBIXCLZ22, DBLP:conf/acl/0009C23}. 
However, recent studies show that they can benefit from training with rationales, by eliciting rationales from large teacher models and using them as additional supervision signals \cite{DBLP:journals/corr/HintonVD15, DBLP:conf/acl/HoSY23, DBLP:conf/acl/HsiehLYNFRKLP23}. 
The most common approach places rationales before the answer, simulating the CoT reasoning process, significantly improving performance on reasoning tasks \cite{DBLP:conf/acl/LiHYRC023, DBLP:conf/icml/FuPOSK23, DBLP:conf/acl/LiHLJSL024}. 
Other methods place rationales after the answer or treat the answer and rationale learning as a multi-task problem, also yielding substantial gains \cite{DBLP:journals/corr/abs-2404-09170, DBLP:conf/emnlp/WadhwaAW24, DBLP:conf/acl/HsiehLYNFRKLP23}. 
Building on them, we construct various rationale-augmented training methods to explore whether rationales can enhance NLU learning, and which is the most effective.

Beyond the trained tasks, rationales could also enhance model generalization to unseen tasks.
Instruction tuning improves LLMs' generalization by training them on diverse tasks and instructions \cite{DBLP:conf/iclr/WeiBZGYLDDL22,DBLP:journals/jmlr/ChungHLZTFL00BW24}.
However, \cite{DBLP:conf/acl/LiHYRC023} shows that with a limited number of tasks, answer-only training causes models to struggle with new tasks, while incorporating rationales mitigates this issue.
Additionally, CoT Collection augments instruction-tuning data with rationales, improving performance on unseen reasoning tasks \cite{DBLP:conf/emnlp/KimJKJYSS23}. 
In this work, we investigate whether rationales could enhance model's generalization on NLU tasks, focusing on both performance and interpretability. 
Since it's labor-intensive to isolate NLU tasks from CoT Collection and align them with real-world tasks, we construct a NLU dataset collection with rationales.

\section{NLU Rationale Collection}
To comprehensively assess the impact of rationales on NLU tasks, we construct a high-quality NLU dataset collection comprising 34 widely-used datasets across 11 tasks. 
Moreover, we develop a rationale evaluation framework and experiment with various prompt designs to generate high-quality rationales. Consequently, we create NLU Rationale Collection with 252,507 rationales. 
Due to limitations in available annotators with domain expertise, the dataset is in Chinese. Below, we outline the dataset construction process.



\subsection{Constructing High-Quality NLU Datasets}\label{NLU_Clean}
We construct a high-quality NLU dataset collection through collecting, clustering and cleaning.

\textbf{Collecting} \  We collect 34 widely used datasets spanning 11 well-studied NLU tasks. 
These tasks are carefully selected to ensure comprehensive coverage of core language understanding capabilities, ranging from low-level lexical classification to high-level causality and commonsense inference.
If original data splits are available, we retain them; otherwise, we divide the dataset into an 8:1:1 ratio for training, validation, and test sets.

\textbf{Clustering} \ Recognizing the importance of data diversity in LLM learning \cite{DBLP:conf/iclr/WeiBZGYLDDL22, DBLP:journals/corr/abs-2306-13840}, we enhance diversity within each dataset. We remove duplicate samples and apply clustering to eliminate semantically redundant instances. Specifically, each sample is encoded using the bge-base-zh-v1.5 model \cite{DBLP:journals/corr/abs-2309-07597}, followed by K-Means clustering \cite{macqueen1967classification} to identify the most representative samples.

\textbf{Cleaning} \ Since label accuracy is crucial for generating high-quality rationales, as identified in our prompt design experiments, we apply a model-human collaborative cleaning to improve dataset quality. Each sample is first evaluated by LLMs-as-a-Judge, an ensemble of three commercial-grade LLMs \cite{DBLP:conf/emnlp/TanLWBJBKL0024}. Samples with labels matching the majority vote are retained, while inconsistent ones are reviewed by college-graduate Chinese-speaking annotators. This process improves dataset quality cost-effectively and mitigates biases from sole reliance on LLMs.
More details on dataset construction and statistics are provided in Appendix A and Appendix E.

\subsection{Generating High-Quality Rationales}\label{NLU_Rat_COL}
We generate rationales for each training sample using Tencent's Hunyuan-turbo\footnote{https://cloud.tencent.com/document/product/1729}, a leading proprietary Chinese LLM. To ensure rationale quality, we establish an evaluation framework for NLU scenarios. Using the framework, we conduct experiments to determine the most effective prompt designs to generate rationales. These rationales are finally refined through rule-based filtering and manual revisions.

\begin{table*}[t]
\centering
\begin{tabular}{p{3cm}p{12cm}}
\hline
\textbf{Dimension}         & \textbf{Definition}                                                                                                                                               \\ \hline
\textbf{Conciseness}                & Rationales avoid unnecessary length or repetition, delivering key points succinctly without losing clarity or comprehension.
\\ \hdashline
\textbf{Comprehensiveness} & Rationales thoroughly address all essential aspects, covering key points and providing necessary context without omissions.                                 \\ \hdashline
\textbf{Logical Coherence} & Rationales follow a clear and human-like logical reasoning process, progressing step-by-step without contradictions or logical gaps.                             \\ \hdashline
\textbf{Faithfulness}      & Rationales accurately reflect the input text and final label, justifying the label based only on the provided evidence.                                   \\ \hline
\textbf{Diversity}         & Rationales are rich in linguistic expressions, reasoning forms and explanation perspectives, avoiding repetitive patterns or reliance on a uniform approach. \\ 
\hline
\end{tabular}
\vspace{-0.2cm}
\caption{Definitions of four quality evaluation dimensions and diversity of rationales.}
\vspace{-0.4cm}
\label{rat_quality_def}
\end{table*}

\textbf{Rationale Evaluation Framework} \  
Evaluation metrics for rationale quality in reasoning tasks are well-established, focusing on logical structures as these tasks rely on rationales to decompose problems and derive solutions \cite{DBLP:conf/iclr/GolovnevaCPCZFC23}.
In contrast, rationales in NLU tasks only need to locate and analyze key information, after which a summary can lead to the answer.
To address the differences, we propose a rationale evaluation framework for NLU tasks.
It defines four core quality dimensions: \textbf{Conciseness}, \textbf{Comprehensiveness}, \textbf{Logical Coherence}, and \textbf{Faithfulness}, with definitions provided in Table \ref{rat_quality_def}.
These dimensions are adapted from common reasoning metrics and align with the characteristics of NLU tasks, reflecting both reasoning aspects and NLU-specific features.
Beyond individual rationale quality, the framework also considers inter-rationale \textbf{Diversity}, considering the critical role of data diversity in LLM learning. 
Additionally, to facilitate human evaluation, it includes detailed 1–5 scoring criteria for each dimension.
Further details on motivation, related metrics, and scoring criteria are provided in Appendix B.1.


\begin{figure}[t]
\centering
\includegraphics[scale = 0.30]{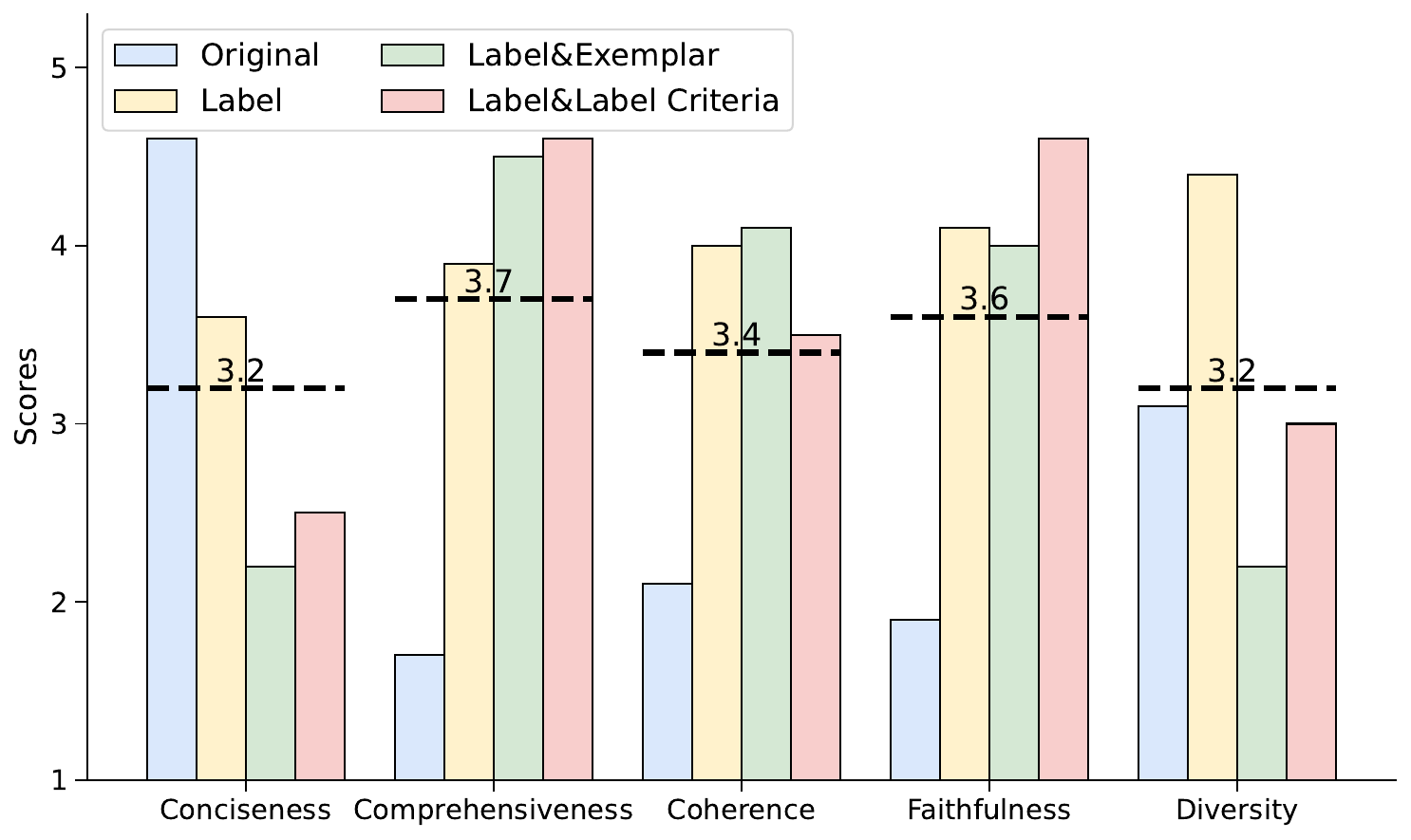}
\caption{Human evaluation on rationale quality of different prompt designs. Black dashed lines indicate average score for each dimension.}
\vspace{-0.4cm}
\label{rat_quality}
\end{figure}

\begin{figure}[t]
\centering
\includegraphics[scale = 0.29]{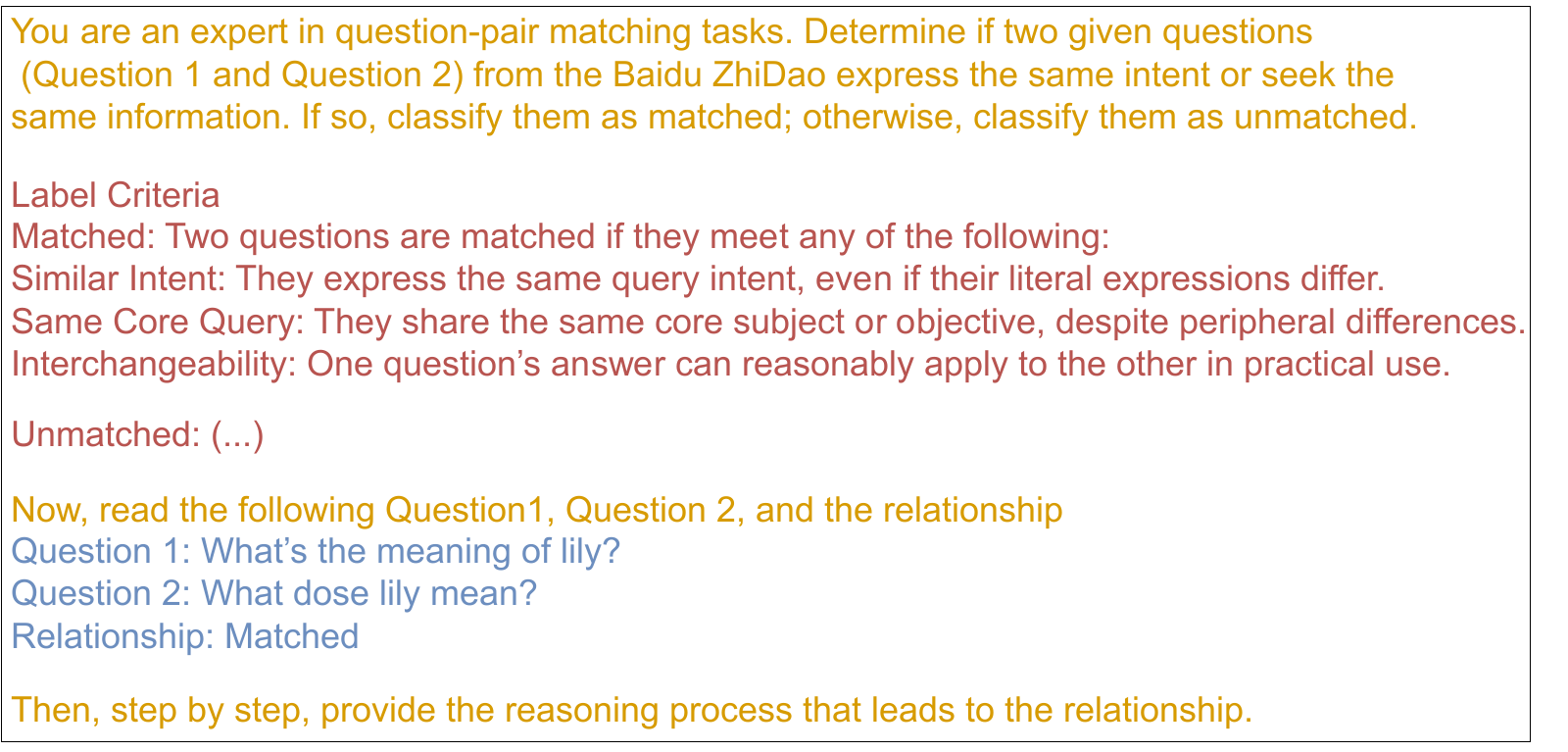}
\vspace{-0.3cm}
\caption{Rationale generation prompt for a Paraphrase Detection task (LCQMC). \textcolor{customorange}{Orange} is the original prompt. \textcolor{customblue}{Blue} is input with the golden label. \textcolor{customred}{Red} is the judge criteria for each label.} 
\vspace{-0.5cm}
\label{fig:prompt}
\end{figure}

\textbf{Prompt Design Process}  \ 
Using the framework, we experiment with different prompts to identify the optimal designs for generating high-quality NLU rationales. Common prompts include the ground truth label and human-written exemplars \cite{DBLP:conf/acl/MagisterMAMS23,DBLP:conf/emnlp/KimJKJYSS23}. Since the fixed label space is specific to NLU tasks, we define judge criteria for each label. 
To evaluate these designs, we randomly sample 100 rationales using different prompt designs and have two annotators score each dimension. The average scores are in Figure \ref{rat_quality}, with specific cases and analysis in Appendix B.2. 
First, simply incorporating the label significantly improves rationale quality, likely because it simplifies the model’s task by focusing on rationale generation rather than solving the task simultaneously. 
And we find that \textbf{label accuracy is crucial}, as incorrect labels lead to redundant and incoherent rationales, underscoring the importance of our label cleaning. 
Additionally, \textbf{exemplars reduce conciseness and diversity}. LLMs can generate high-quality rationales in a zero-shot setting, whereas exemplars restrict the reasoning styles and perspectives. Moreover, \textbf{label criteria trade diversity for comprehensiveness and faithfulness}. Based on these insights, we use zero-shot prompts with golden labels to guide LLMs in generating rationales, as shown in Figure \ref{fig:prompt}, while some incorporate label criteria, improving quality without sacrificing diversity.

Despite our efforts in data cleaning and prompt designs, we still observe some rationales with incorrect final answers.
To mitigate this issue, we apply rule-based filtering and manual revisions to each rationale, as detailed in Appendix B.3.

\section{Rationale-Augmented Methods}
\begin{figure}[t]
\centering
\includegraphics[scale = 0.65]{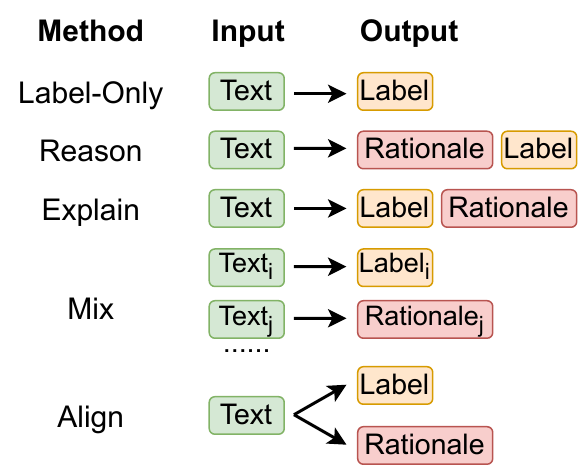}
\caption{Examples of training methods}.%
\vspace{-0.4cm}
\label{method}
\end{figure}

To empirically evaluate the impact of rationales on NLU tasks, we construct and compare various rationale-augmented training and inference methods in NLU scenarios, as shown in Figure \ref{method}.
These methods are based on existing work, including those that have been widely proven to improve reasoning tasks.
Let the model be denoted as $f$, and a training sample $(x_i, y_i, r_i)$ represents the input text, the label, and the rationale, respectively.

In training stage, the most basic approach  $f(x_i) \rightarrow y_i$ is denoted as \textbf{Label-Only}, where only the label is learned \cite{DBLP:conf/iclr/WeiBZGYLDDL22}. To integrate rationales, most works append them before the labels $f(x_i) \rightarrow (r_i, y_i)$, which we term  \textbf{Reason} \cite{DBLP:conf/acl/HoSY23,DBLP:conf/acl/LiHYRC023, DBLP:conf/icml/FuPOSK23, DBLP:conf/acl/LiHLJSL024}. 
Furthermore, \cite{DBLP:journals/corr/abs-2404-09170,DBLP:conf/emnlp/WadhwaAW24} have demonstrated that performance improvements could also be achieved by placing rationales after the labels $f(x_i) \rightarrow (y_i, r_i)$, which we term \textbf{Explain}.
Beyond these in-sample combinations, inspired by \cite{DBLP:conf/acl/HsiehLYNFRKLP23}, which treats label and rationale learning as a multi-task learning problem, we propose inter-sample combination methods.
In \textbf{Mix}, we mix $f(x_i)\rightarrow y_i$ and $f(x_i)\rightarrow r_i$ samples, randomly drawing $N$ samples from them to form batches, with the loss defined as:
\begin{equation*}
\begin{aligned}
\mathcal{L}_{\text{Mix}} = \frac{1}{\sum_{i=1}^{N} T_i} \sum_{i=1}^{N} \sum_{t=1}^{T_i} \mathcal{L}_{it},
\mathcal{L}_{it} =
\begin{cases} 
\ell (f(x_i), y_{it}) \\
\ell (f(x_i), r_{it}),
\end{cases}
\label{loss_mix}
\end{aligned}
\end{equation*}
where $T$ denotes token number in the label or rationale, and $\ell$ is cross-entropy loss per token. 
However, common loss calculation averages the losses across all tokens in the batch. 
This means that in the above methods, long rationales would overwhelm labels, leading to a low loss and insufficient learning for labels.
To address the issue, we further develop \textbf{Align}, which separately calculates losses for labels and rationales within a batch. Align binds $f(x_i) \rightarrow y_i$ and $f(x_i) \rightarrow r_i$ from the same instance into a single training batch, computes losses individually then sums, as below:
\begin{equation*}
\begin{aligned}
\mathcal{L}_{\text{Align}} =
\begin{array}{c}
  \frac{1}{\sum_{i=1}^{N} T_i} \sum_{i=1}^{N} \sum_{t=1}^{T_i} \ell (f(x_i), y_{it}) \\
  + \frac{1}{\sum_{i=1}^{N} T_i'} \sum_{i=1}^{N} \sum_{t=1}^{T_i'} \ell (f(x_i), r_{it}).
\end{array}
\label{loss_align}
\end{aligned}
\end{equation*}

In the inference stage, we consider three zero-shot methods: directly generating the label $f(x_i) \rightarrow y_i$ as \textbf{Direct}, generating the rationale followed by the label $f(x_i) \rightarrow (r_i, y_i)$ as \textbf{CoT}, and generating the label followed by the rationale $f(x_i) \rightarrow (y_i, r_i)$ as \textbf{Rationalize}.
Since both Direct and Rationalize output the label first, we omit Rationalize inference during performance evaluation.

\section{Exploration Experiments}
In this section, we introduce three research questions, systematically exploring the impact of rationales on NLU in terms of inference, training, and unseen tasks. To empirically answer these questions, we conduct extensive experiments on the NLURC dataset, comparing and analyzing various rationale-augmented methods.

\textbf{Research Questions} \ 
Incorporating rationales, either generating them before answering during inference or placing them around the original answers during training, significantly improves LLMs performance on complex reasoning tasks \cite{DBLP:conf/nips/KojimaGRMI22,DBLP:conf/acl/LiHLJSL024}. 
In contrast to these reasoning tasks, NLU tasks often require only the understanding of key information to obtain correct answers.
This raises the question: Can the integration of rationales into inference and training similarly enhance NLU performance? 
In this work, we investigate the following research questions about the effectiveness of rationales in NLU tasks. 
(\textbf{RQ1}) Does generating rationales before providing answers during inference benefit NLU tasks? 
(\textbf{RQ2}) Does incorporating rationales into training improve NLU performance over label-only training, and if so, what is the most effective approach?
(\textbf{RQ3}) Does rationale-augmented training enhance the model’s generalization ability on unseen tasks?


\textbf{Experimental Setup} \ 
We utilize the Qwen1.5-Chat model series \cite{DBLP:journals/corr/abs-2309-16609}, covering model sizes of 0.5B, 1.8B, 7B and 32B. 
All training and evaluation data come from our NLURC dataset, with both conducted at the task level using all datasets within each task. 
We have two training scenarios: single-task training on the evaluation task (RQ1, RQ2), reporting average results from three runs with different random seeds; and multi-task  training on all tasks except the evaluation task (RQ3), with a single run. 
Evaluation is performed on the test sets of each dataset in the evaluation task, with macro-average performance reported as the final result.
We select five diverse tasks for evaluation, with increasing average rationale length across datasets in each task: \textbf{Reading Comprehension} (Read. Comp.), \textbf{Stance Detection} (Stance), \textbf{Topic Classification} (Topic), \textbf{Chinese Linguistics} (Linguistics), and \textbf{Reading Comprehension with Common Sense} (RC. /w CS.). See Appendix D.1 for further details.

\begin{table*}[t]
\centering
\begin{tabular}{lllcccccc}
\hline
\textbf{Sizes} &\textbf{Infer} &\textbf{Train} & \textbf{Read. Comp.} & \textbf{Stance} & \textbf{Topic} & \textbf{Linguistic} & \textbf{RC. w/CS.} & \textbf{Average} \\ \hline
\multirow{10}{*}{0.5B}  & \multirow{4}{*}{CoT}   & Original & 16.76 & 37.95 & 23.96 & 5.60 & 35.46 & 23.95 \\             
&& Reason    & 66.78       & 73.92  & 78.77 & 45.51   & 41.28    & 61.25 \\ 
&& Mix      & 69.72       & 77.31  & 83.24 & 49.05   & 41.33    & 64.13 \\ 
&& Align     & 72.11       & 78.47  & 85.23 & 47.79   & 43.89    & 65.50 \\ \cdashline{2-9}

& \multirow{6}{*}{Direct} & Original & 14.99 & 30.35 & 26.86 & 7.69 & 40.11 & 24.00 \\
&& Label-Only      & \underline{77.80}        &\underline{83.20}   & \underline{91.55} & \underline{57.77}   & \underline{56.61}    & \underline{73.39} \\ 
&& Explain      & 71.78       & 78.78  & 86.75 & 52.55   & 50.57    & 68.09 \\ 
&& Mix      & 75.22       & 82.02  & 89.31 & 54.79   & 51.03    & 70.47 \\
&& \textbf{Align}     & \textbf{79.47}       & \textbf{86.11}  & \textbf{92.08} & \textbf{64.69}   & \textbf{62.78}    & \textbf{77.03} \\ \hline

\multirow{10}{*}{7B}     & \multirow{4}{*}{CoT}   & Original & 71.91 & 71.23 & 71.36 & 42.95 & 62.89 & 64.07 \\ 

&& Reason      & 84.87      & 89.32  & 91.42 & 66.18   & 68.33    & 80.02 \\ 
&& Mix      & 86.61       & 90.75  & 92.20 & 68.19   & 72.07    & 81.96 \\ 
&& Align     & 85.26       & 90.78  & 91.43 & 67.12   & 74.45    & 81.81 \\ \cdashline{2-9}

& \multirow{6}{*}{Direct} & Original & 72.22 & 71.47 & 66.31 & 37.01 & 67.32 & 62.87 \\
&& Label-Only      & \underline{87.46}       & \underline{94.41}  & \underline{94.54} & \underline{76.68}   & \underline{89.45}    & \underline{88.51} \\ 
&& Explain      & 85.20        & 91.16  & 94.15 & 74.25   & 77.32    & 84.42 \\ 
&& Mix      & 87.18       & 93.84  & 94.39 & 76.66   & 82.85    & 86.98 \\ 
&& \textbf{Align}     & \textbf{87.72}       & \textbf{94.51}  & \textbf{94.73} & \textbf{76.92}   & \textbf{90.17}    & \textbf{88.81} \\ \hline

\multirow{10}{*}{32B}  & \multirow{4}{*}{CoT}    & Original & 80.22 & 88.88 & 85.55 & 65.82 & 80.91 & 80.28 \\ 
&& Reason    & 87.50        & 94.25  & 93.64 & 74.15   & 85.63    & 87.03 \\
&& Mix      & 88.37       & 94.37  & 93.97 & 78.30   & 94.67    & 89.94 \\ 
&& Align     & 88.42       & 94.49  & 93.78 & 74.66   & 89.06    & 88.08 \\ \cdashline{2-9}

& \multirow{6}{*}{Direct} & Original & 78.34 & 83.31 & 82.70 & 63.63 & 84.21 & 78.44 \\ 
&& Label-Only      & \underline{90.41}       & 95.93  & \underline{95.54} & 80.63   & \underline{94.40}     & \underline{91.38} \\  
&& Explain      & 89.33      & 94.88  &94.94 & 79.56   & 90.70     & 89.88 \\ 
&& Mix      & 90.15       & \underline{95.94}  & 95.15 & \underline{80.84}   & 92.86    & 90.99 \\
&& \textbf{Align}     & \textbf{90.52}       & \textbf{96.58}  & \textbf{95.81} & \textbf{81.30}    & \textbf{94.67}    & \textbf{91.78} \\ \hline
\end{tabular}
\caption{Results of different methods on seen tasks. Average rationale length of the task rises from left to right. The best comparable
performances are \textbf{bold} and second best \underline{underlined}.}
\label{results_all}
\end{table*}

\subsection{RQ1: Inference with Rationales}
To assess the effect of rationales during inference, we compare Direct and CoT inference across five evaluation tasks, as shown in Table~\ref{results_all} and Appendix Table 14. 
For each setting, we report both the original model performance and after single-task training. Since models trained with Label-Only and Explain can only produce labels, we omit their CoT inference results; conversely, Reason-trained models may not directly output answers, so we omit their Direct inference results.

\vspace{0.1cm}
\noindent \textit{Does CoT inference benefit NLU tasks?} 

\textbf{The impact depends on the model size, with a positive correlation.}
For the original model, the absolute difference in average performance between CoT and Direct inference for the 0.5B and 1.8B models is -0.05, suggesting that CoT inference slightly harms small models. But as the model size increases, it begins to benefit NLU performance and the gains increase, with the difference for the 7B and 32B models being +1.2 and +1.84, respectively. 
This indicates that the effect of CoT inference on NLU tasks is model-size dependent and positively correlated.
A similar trend is observed in rationale-trained models, whose CoT performance largely surpasses that of the original models.
Take Align-trained models as examples, the average performance differences between CoT and Direct inference from 0.5B to 32B are from -11.53 to -3.7.
Although generating rationales consistently impairs performance, this negative impact decreases as model size grows. Due to computational limits, we train models up to 32B. But this trend suggests that, for sufficiently large trained models, CoT inference could be beneficial, consistent with findings that CoT boosts performance on reasoning tasks only when applied to large models \cite{DBLP:conf/nips/Wei0SBIXCLZ22}.
Thus, CoT inference can benefit NLU tasks, depending on model size.


\begin{figure}[t]
\centering
\includegraphics[scale = 0.328]{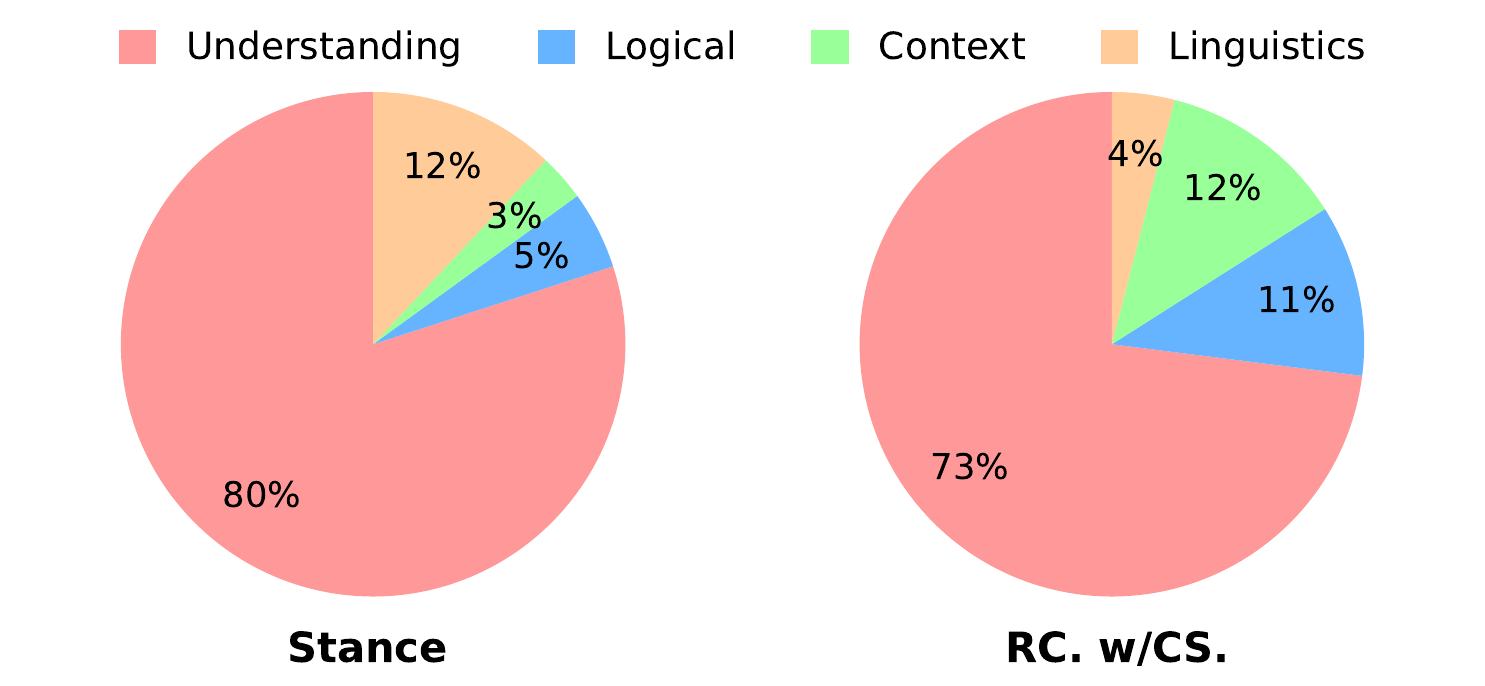}
\vspace{-0.2cm}
\caption{Align-trained 7B model's CoT error type distribution in Stance and RC. w/CS. task.} 
\vspace{-0.4cm}
\label{error_pie}
\end{figure}

\vspace{0.1cm}
\noindent \textit{What leads to the negative impact of CoT inference on NLU performance?} 

\textbf{CoT inference tends to analyze the entire text, increasing the chance of understanding errors, especially for smaller models with weaker understanding abilities.}
One might intuitively guess the negative impact of CoT inference stems from the limited reasoning capacity of small models. To understand why CoT errors occur, we analyze cases from Align-trained 7B models where CoT fails but Direct inference succeeds. We sample 200 examples from each of the Stance and RC. w/CS. tasks and have graduate-level annotators identify the error types.
Based on the CoT error analysis in \cite{DBLP:conf/acl/TongLWWTS24}, we categorize CoT errors in NLU tasks into four types: Understanding, Logical, Context, and Linguistic. See Appendix Tables 12 and 13 for definitions and examples.
Surprisingly, most CoT errors are about understanding rather than logic, as shown in Figure \ref{error_pie}.
Through further case analysis, we find that CoT reasoning for NLU tasks focuses on analyzing different sections of the text, rather than solving the problem step by step. 
These reasoning processes are logically consistent, but errors arise when specific parts of the text are misanalyzed, which disrupt the subsequent reasoning chain and lead to incorrect summaries.

But why do understanding errors in CoT reasoning harm NLU performance more than in Direct inference?
In NLU tasks, analyzing key parts of the input is often sufficient to solve the problem.
\textbf{But reasoning entails a more thorough analysis of the entire input, increasing the chance of understanding errors.} 
For original models, on tasks with short rationales, like Read. Comp. or Stance, CoT inference performs comparably or better than Direct inference.
But on tasks with long rationales, such as RC. w/CS., CoT consistently harms performance. 
As for trained models, this negative impact of CoT is more pronounced across different tasks. 
Despite improvements in understanding abilities, reflected in better CoT and Direct inference performance, training mainly enhances the understanding on key input components, leaving the overall text analysis ability less enhanced and widening the gap. 
These findings indicate that CoT inference requires more comprehensive text analysis, increasing the chance of errors, especially for smaller models with weaker understanding abilities.

\subsection{RQ2: Training with Rationales}
We now examine the impact of rationales on training, focusing on Direct inference, as RQ1 shows CoT consistently underperforms on trained models.

\vspace{0.1cm}
\noindent \textit{Does training with rationale improve performance over label-only training?}

\textbf{Yes, but only one method achieves consistent improvements.}
We observe that Align-trained models consistently outperform others across tasks and model sizes, indicating that learning rationales can enhance NLU performance.
Notably, Align is the only method that consistently outperforms Label-Only training. In contrast, Mix shows marginal gains only with the 32B model, and the widely used Reason and Explain methods, despite their success in reasoning tasks \cite{DBLP:conf/acl/HoSY23, DBLP:conf/emnlp/WadhwaAW24}, consistently underperform.
This suggests that while rationales contain valuable information, only certain training strategies can effectively harness them.

\begin{figure}[t]
\centering
\includegraphics[scale = 0.40]{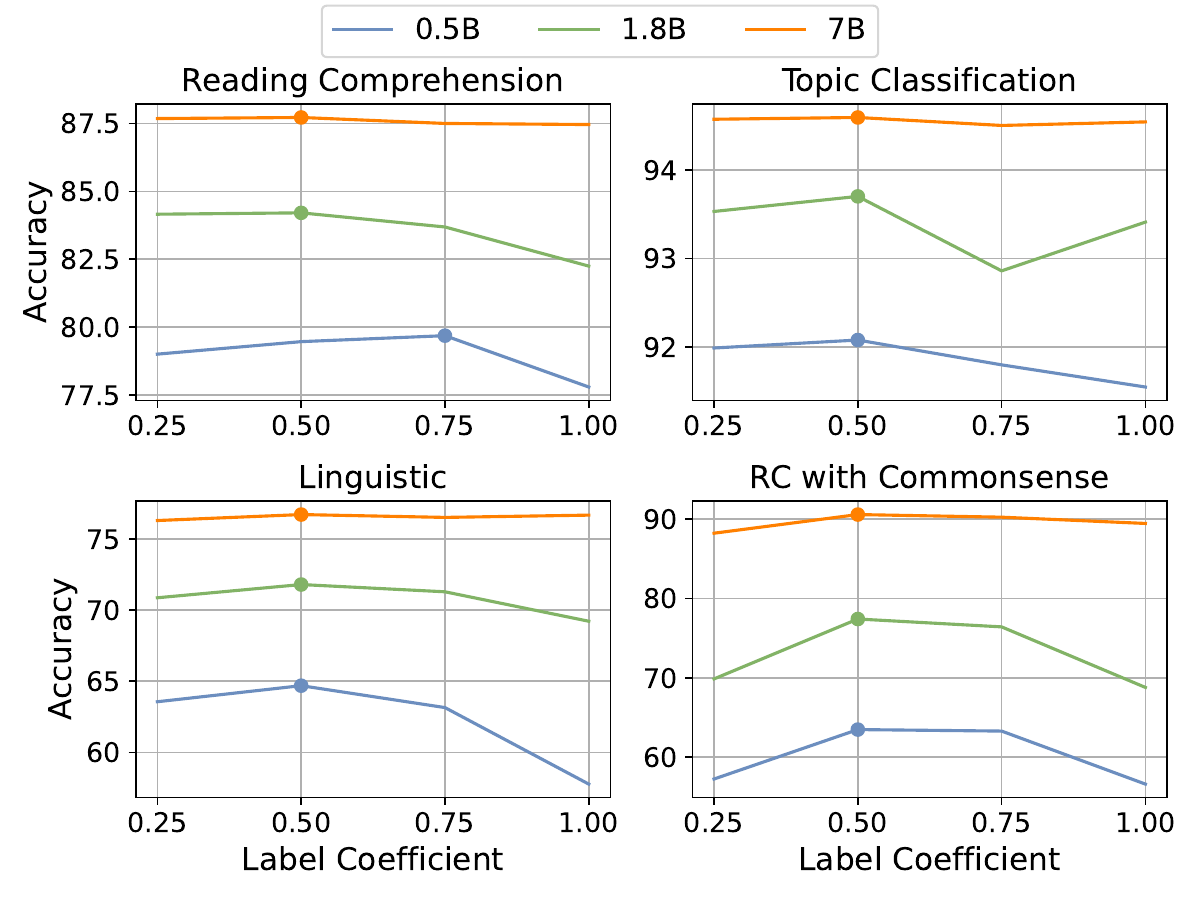}
\caption{Performance of varying label and rationale coefficients in Align training. The X-axis values represent the label coefficient, and 1 corresponds to Label-Only training.}
\vspace{-0.3cm}
\label{fig_cof}
\end{figure}

\begin{figure}[t]
\centering
\includegraphics[scale = 0.22]{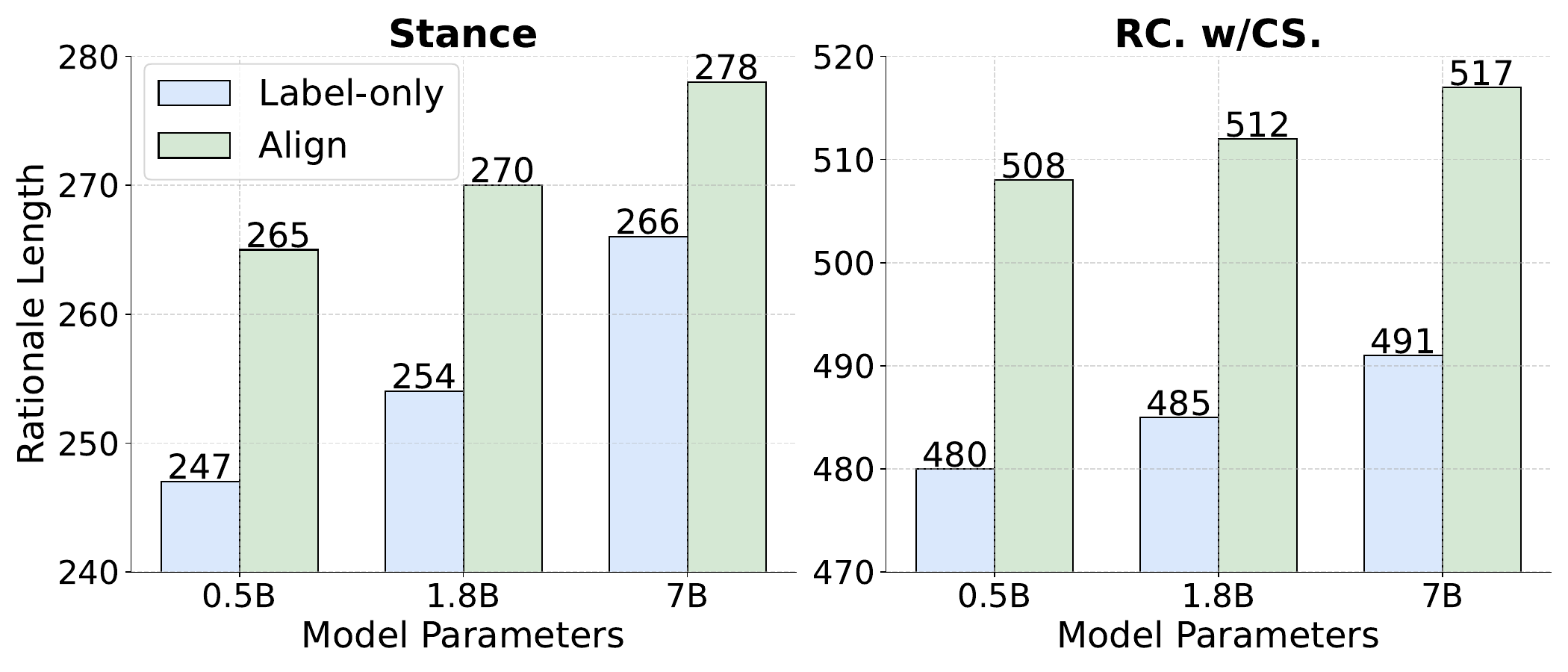}
\caption{Average rationale length of correctly answered samples by models trained with Align and Label-Only.} 
\vspace{-0.3cm}
\label{ratlen_comp}
\end{figure}

\vspace{0.1cm}
\noindent \textit{Why does only Align achieve consistent gains?}

\textbf{Preserving label learning is critical when incorporating rationales.}
The main distinction between Align and other rationale training methods lies in its emphasis on preserving label learning by separate loss calculations.
As identified in RQ1, while CoT's comprehensive text analysis may aid in solving NLU tasks, the key challenge remains in understanding the crucial parts of input.
By balancing label and rationale learning, Align not only captures rationale information, but also learns the mapping from texts to labels, enhancing model's ability to comprehend key aspects of the input.
In Figure \ref{fig_cof}, we present an ablation study of Align’s loss coefficient, with detailed results in Appendix D.4.
The results show that, for different models and tasks, incorporating rationales consistently outperforms label-only training. 
Notably, the best performance is achieved when both the label and rationale coefficients are set to 0.5, underscoring the importance of balanced learning of rationale and label.
Thus, while learning rationales could improve NLU performance, it is crucial to maintain a strong focus on label learning.

\textbf{Align training enables models to tackle more informative samples.}
To better understand the performance gains of Align over Label-Only training, we examine cases where Align succeeds but Label-Only fails. We find that these samples are more informative and require longer rationale to analyze. 
To quantify this, we compare the rationale lengths for samples correctly predicted by models trained with Align and Label-Only, respectively. We generate rationales for each set using Hunyuan-Turbo and calculate the average lengths. The results in Figure \ref{ratlen_comp} show that rationales for samples correctly handled by Align are consistently longer than those by Label-Only across tasks and model sizes. This suggests that Align better equips models to handle examples with abundant information.

\begin{table*}[t]
\centering
\begin{tabular}{lcccccccccccc}
\hline
\textbf{Task} & \multicolumn{3}{c}{\textbf{0.5B}} & & \multicolumn{3}{c}{\textbf{1.8B}} & &  \multicolumn{3}{c}{\textbf{7B}} & \textbf{72B} \\ \cdashline{2-4} \cdashline{6-8} \cdashline{10-12}
                 & \textbf{Ori} & \textbf{Label} & \textbf{Align} & & \textbf{Ori} & \textbf{Label} & \textbf{Align} & & \textbf{Ori} & \textbf{Label} & \textbf{Align} & \textbf{Ori} \\ \cline{2-4} \cline{6-8} \cline{10-12}
 Read. Comp. & 15.0 & 28.2 & \textbf{39.4} & & 39.2 & 63.9 & \textbf{68.0}& & 72.2 & 80.8 & \textbf{81.4} & 80.0 \\ 
           & & \scalebox{1}{(+88\%)} & \scalebox{1}{(\textbf{+163\%})} & & & \scalebox{1}{(+63\%)} & \scalebox{1}{(\textbf{+73\%})} & & & \scalebox{1}{(+12\%)} & \scalebox{1}{(\textbf{+13\%})} & \\ 
Stance        & 30.3 & 44.5 & \textbf{51.0} & & 46.7 & 67.8 & \textbf{74.3} & & 71.5 & 87.0 & \textbf{89.4} & 89.2 \\ 
           & & \scalebox{1}{(+47\%)} & \scalebox{1}{(\textbf{+68\%})} & & & \scalebox{1}{(+45\%)} & \scalebox{1}{(\textbf{+59\%})} & & & \scalebox{1}{(+22\%)} & \scalebox{1}{(\textbf{+25\%})} & \\ 
Linguistic    & 7.7 & 26.9 & \textbf{41.1} & & 23.2 & 43.5 & \textbf{47.1} & & 37.0 & 52.4 & \textbf{53.2} & 58.2 \\
           & & \scalebox{1}{(+249\%)} & \scalebox{1}{(\textbf{+434\%})} & & & \scalebox{1}{(+88\%)} & \scalebox{1}{(\textbf{+103\%})} & & & \scalebox{1}{(+42\%)} & \scalebox{1}{(\textbf{+44\%})}  & \\ 
RC. w/CS.    & 40.1 & 43.6 & \textbf{44.2} & & 53.6 & 56.7 & \textbf{59.2} & & 67.3 & 72.6 & \textbf{73.1} & 84.6 \\
           & & \scalebox{1}{(+9\%)} & \scalebox{1}{(\textbf{+10\%})} & & & \scalebox{1}{(+4\%)} & \scalebox{1}{(\textbf{+10\%})} & & & \scalebox{1}{(+8\%)} & \scalebox{1}{(\textbf{+9\%})} & \\ 
\hline
\end{tabular}
\vspace{-0.2cm}
\caption{Results on unseen tasks. Ori and Label denote the original and Label-Only-trained models, respectively.}
\label{unseen_performance}
\end{table*}


\subsection{RQ3: Impacts on Unseen Tasks } \label{experiment_tuning}
In this section, we investigate the impacts of rationale training on model generalization, specifically comparing it to Label-Only training.
Unlike RQ1 and RQ2, where training and evaluation are conducted on the same task, we now evaluate models on unseen tasks.
For each evaluation task, models are trained on the full NLURC dataset, excluding data from the specific task being evaluated.
In addition to performance, we assess interpretability bby evaluating the quality of the model's rationales. NLU tasks are widely used in domains that require transparency, and rationales serve as a key indicator of a model’s decision-making process \cite{DBLP:journals/corr/abs-2303-17564,DBLP:journals/frai/GurrapuKHLB23}.
Based on prior findings, we use Align training with Direct inference for performance evaluation and Rationalize inference for rationale evaluation.


\textbf{Rationales improve models' generalization performance.}
Table \ref{unseen_performance} compares the Direct inference performance between the original, label-only trained and rationale-trained models.
The results show that label-only training already leads to performance improvements.
However, incorporating rationales into the training process further substantially amplifies these gains, particularly for smaller models.
Notably, the Align-trained 7B model outperforms the 72B model on both the Read. Comp. and Stance tasks, despite the latter being ten times larger in size.
This suggests that the improvements driven by rationales go beyond the benefits of multi-task learning, highlighting the critical role of rationales in enhancing NLU performance.

\begin{figure}[t]
\centering
\includegraphics[scale = 0.4]{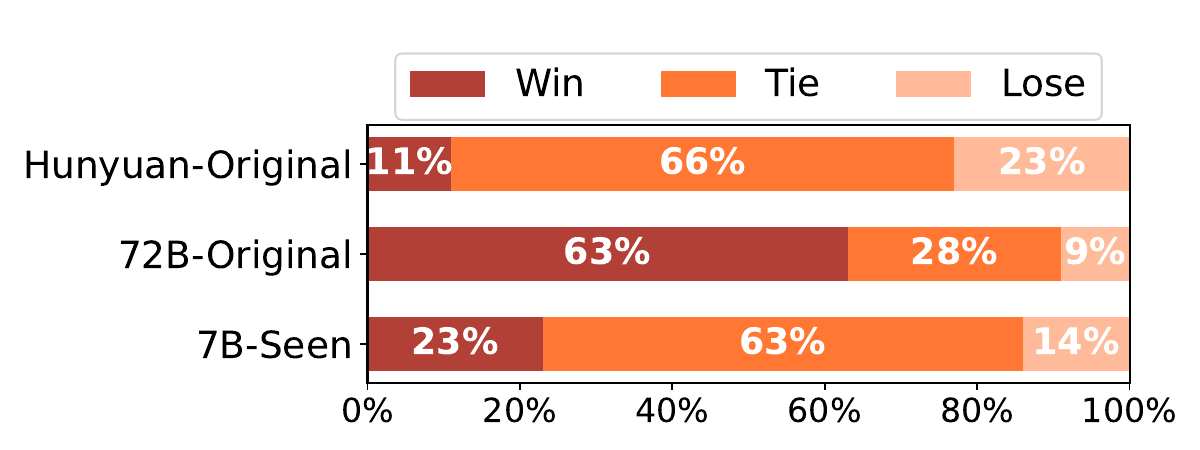}
\caption{Human evaluation of rationale quality, comparing the 7B-Unseen model with the Hunyuan-Original, 72B-Original, and 7B-Seen models.}
\vspace{-0.3cm}
\label{unseen_quality}
\end{figure}

\textbf{Rationales improve models' generalization interpretability.}
We compare the quality of rationales generated by the Qwen-7B model trained on NLURC data excluding the evaluation task (7B-Unseen), with that of original Hunyuan-turbo (Hunyuan-Original), original Qwen-72B (72B-Original), and Qwen-7B trained solely on the evaluation task (7B-Seen).
For each of the four evaluation tasks, we randomly sample 50 rationales from each model using Rationalize inference and have college-graduate annotators to make win, tie, or lose judgments based on our proposed evaluation dimensions, excluding Diversity.
The results are shown in Figure \ref{unseen_quality}, with detailed case analyses in Appendix D.5.
It demonstrates that the 7B-Unseen model produces rationales of comparable quality to Hunyuan-Original, a leading proprietary LLM, and even outperforms the 72B-Original, despite it being ten times larger. 
These findings underscore the effectiveness of rationales in enhancing the quality of rationales generated by LLMs, as well as their interpretability.
Furthermore, the 7B-Unseen outperforms 7B-Seen, suggesting that rationale diversity plays a key role in improving the model's rationalizing capabilities.
In conclusion, using our NLURC dataset for instruction tuning not only enhances the model’s generalization on NLU tasks but also improves its interpretability.

\section{Conclusion}
This paper investigates whether, how and why rationale-augmented training and inference impact NLU tasks, uncovering three key findings: (1) Impact of CoT inference shifts from negative to positive as model size grows, as reasoning tends to over-analyze the text and leads to higher chances of understanding errors in smaller models; (2) Align training consistently improves performance, indicating the crucial role of label learning in NLU tasks; (3) Integrating rationales into multi-task training enhances both performance and interpretability on unseen NLU tasks. Additionally, we release NLURC, a comprehensive NLU dataset collection with rationales. It serves as a valuable resource for both exploration and instruction-tuning data, aiming to improve LLM generalization and interpretability in NLU tasks.

\bibliography{aaai2026}

\clearpage 
\section*{Appendix}
\appendix

\begin{figure*}[t]
\centering
\includegraphics[scale = 0.4]{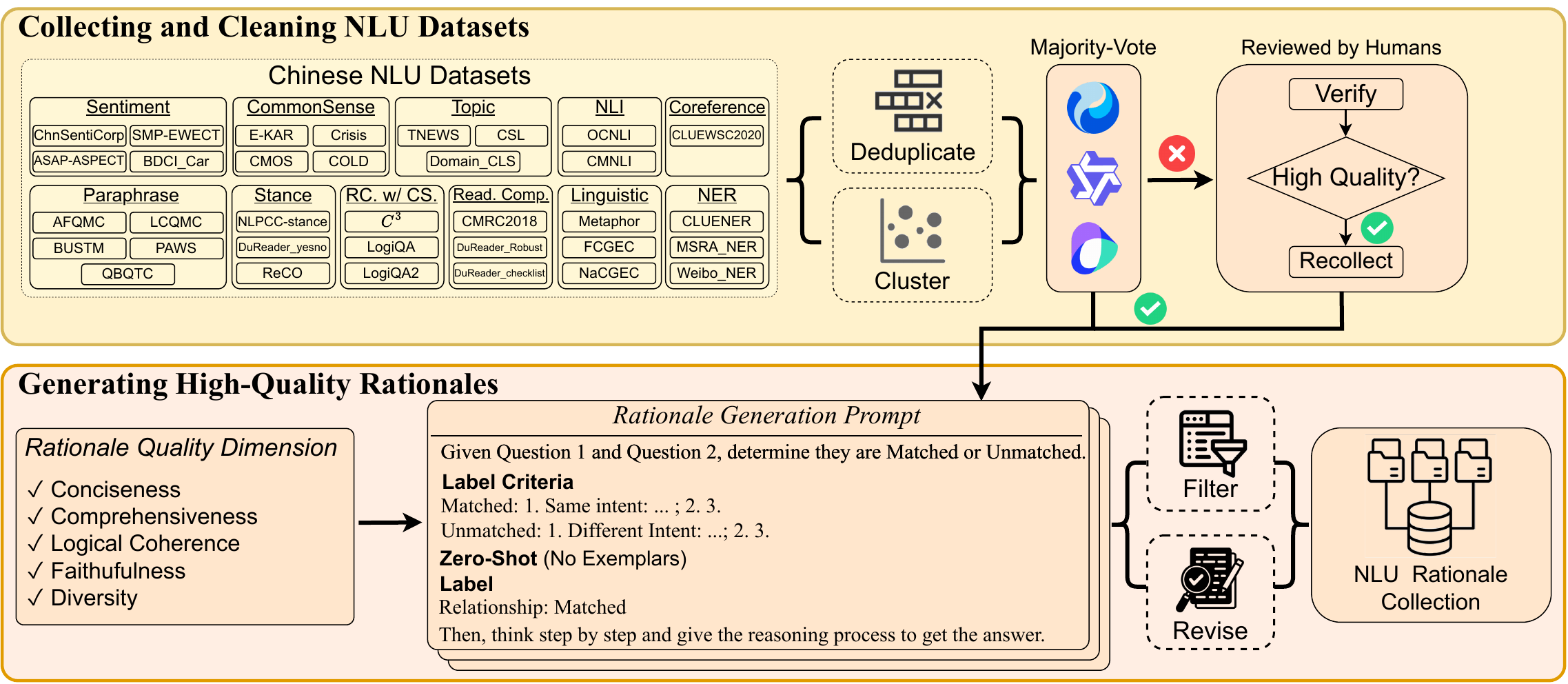}
\caption{The construction process of NLU Rationale Collection.} %
\label{dataset}
\end{figure*}

\section{NLU Collection Construction}\label{Appendix_NLU_Collection}
We present the complete process of constructing the NLU Rationale Collection in Figure \ref{dataset}.
In this section, we describe the process of collecting and cleaning NLU datasets.
We first describe the process of dividing the datasets into training, validation, and test sets. 
We then provide additional information on data cleaning. 
Specifically, we explain the settings in LLMs-as-a-Judge. Next, we outline the operations in the further human reviews. Finally, we analyze the effects of data quantity and quality on LLMs' learning NLU tasks. Detailed information about each dataset is in Appendix \ref{Appendix_datasets} and Appendix Table \ref{app:dataset_combined}.

\subsection{Training, Validation, and Test Split}
For each collected dataset, we retain the original split where available; otherwise, we divide the dataset into an 8:1:1 ratio for training, validation, and test sets. 
Regarding clustering, each training set contains no more than 25,000 samples, while the validation and test sets do not exceed one-eighth of the training set's size. Ultimately, our NLU dataset collection comprises high-quality and diverse training, validation, and test sets, which not only enable effective training but also facilitate model evaluation.

\subsection{LLMs-as-a-Judge}
We utilize LLMs-as-a-Judge to perform an initial round of label accuracy refinement. Specifically, we select three of the most advanced closed-source LLMs:  Hunyuan-turbo\footnote{https://cloud.tencent.com/document/product/1729}, Doubao-pro-4k \footnote{https://www.volcengine.com/product/doubao}, and Qwen-turbo \footnote{https://help.aliyun.com/zh/model-studio/developer-reference/what-is-qwen-llm}.
To balance the calling cost and accuracy, we adopt an 8-shot prompting approach when calling the APIs, using a temperature of 0. Instead of having the LLMs directly evaluate the correctness of the original labels, we instruct them to output the correct label. The final prediction from the Judge was determined by majority vote. In cases where all three LLMs produce different answers, we default to the prediction from Hunyuan-turbo.

\subsection{Human Review} \label{details_in_human_analysis}
After the initial filtering using LLMs, we retain the consistent samples. To ensure that the distribution of the final collection is as close as possible to the original, thus preventing bias introduced by the solely use of LLMs, we conduct a human review for inconsistent samples. We randomly sample 500 samples from the remaining set and ask a human annotator to judge whether their original label is correct, wrong, or ambiguous, which lacks a suitable label, Datasets with more than 50\% correctness are considered high quality, and we will recollect the remaining samples. Specifically, we cluster 20\% of the samples and then verify the correctness of the labels through human analysis. With this second round of human review, we maintain the distribution of the dataset after cleaning at a relatively low cost.

\section{Rationale Generation}\label{Appendix_rationale_generation}
In this section, we provide a detailed explanation of the rationale generation process, as illustrated in Figure \ref{dataset}."
We first present the specific motivation and scoring criteria for each quality dimension. Then, we provide different prompt designs and the corresponding generated rationales.
After that, we provide the rules for filtering rationales, and demonstrate the overall diversity of our rationale collection.

\subsection{Motivation for Rationale Quality Dimensions}\label{Appendix_motivation_rat_quality}
In this subsection, we illustrate the reason and motivation behind our proposed quality dimensions for evaluating rationales. There has been extensive work on assessing the quality of step-by-step reasoning in complex tasks \cite{DBLP:conf/iclr/GolovnevaCPCZFC23,DBLP:conf/emnlp/ChenCSS24}. Although these evaluations encompass a wide range of tasks and metrics, the differences in application scenarios prevent their direct applicability to our context. Given the uniqueness of our scenario, we define a set of clear evaluation dimensions, with specific scoring criteria from 1 to 5 for each dimension detailed in Table \ref{app:scoring_criteria}. Next, we will analyze these dimensions individually, discussing their connections and differences with commonly used dimensions, and clarify the reasons for using them. 

\textbf{Conciseness} Previous works often propose similar dimensions, such as Repetition or Redundancy, to evaluate whether the steps are repetitive or superfluous. However, the nature of NLU tasks means that the logical chains of rationales are usually short. Thus, the core issue with NLU rationales isn’t redundant reasoning steps, but rather useless redundancy in each step, such as repetitive phrasing, inclusion of irrelevant details, excessive elaboration on well-understood concepts, verbose expressions, and linguistic filler. These elements are assessed under the dimension of Conciseness to ensure that rationales are direct and efficiently convey the necessary information.

\textbf{Comprehensiveness} Previous works often propose dimensions like Informativeness and Semantic Coverage to evaluate whether the overall reasoning chain is complete enough to reach the final answer. In NLU tasks, however, our focus is not on the amount of information in each step, but whether it verifies the final answer from various angles. This includes assessing if the rationale addresses all potential counterarguments, incorporates relevant contextual information, and links each piece logically to construct a holistic understanding. Overall, we care about whether the rationale covers all aspects relevant to the task, which we evaluate with Comprehensiveness to ensure it encompasses the breadth  necessary for thorough rationales.

\textbf{Logical Coherence} Previous works propose similar dimensions to evaluate whether the reasoning steps are consistent with themselves and check for logical fallacies. Although we also focus on the logical consistency between steps, the nature of NLU tasks only requires us to ensure that rationales have the logical progression without logical errors.

\textbf{Faithfulness} Similar to previous works, we use this dimension to evaluate whether rationales accurately reflect the model's decision-making process or if there's any misunderstanding of the provided statement. In addition, we assess whether the rationale is based on the provided statement, which in prior works is often judged by Hallucination.

\textbf{Diversity} This dimension has not been evaluated in previous works. Our rationales not only provide interpretations for the model's individual outputs but also serve as instructional data to train models. Considering the emphasis on data diversity during LLMs' learning \cite{DBLP:conf/nips/ZhouLX0SMMEYYZG23}, we encourage diversity in wording, perspective, and structural presentation of rationales to better train LLMs.

\subsection{Prompt Design Experiment}\label{Appendix_prompt_design_experiment}

Based on the evaluation framework for rationales quality that we propose, we conduct prior experiments to identify the best prompt designs in NLU scenarios. We explore the impact of three designs: ground truth labels, few-shot exemplars, and label criteria on the quality of rationales. For different prompt designs, we randomly select 50 generated rationales from datasets ASAP-ASPECT (Sentiment), LCQMC (Paraphrase), NLPCC-stance (Stance), and $C^3$ (RC. w/CS.). Two human annotators score each rationale across various dimensions and take the average. When generating rationales, the sampling temperature is set to 0.7

First, we focus on the original prompt with only the task instruction and sample input, which are shown in Table \ref{prompt_rat_rationale_original}. It can be observed that \textbf{the rationales are very brief, seemingly aimed at providing quick responses, primarily analyzing from one angle with only one logical reasoning step}. Therefore, its Comprehensiveness and Logical Coherence are low, with medium Diversity between samples. Additionally, without the guidance of ground truth labels, it may produce erroneous reasoning and final answers, hence its Faithfulness is very low.

Next, we explore the impact of adding ground truth labels in Table \ref{prompt_rat_rationale_label}. \textbf{Simply adding a label can significantly enhance the quality of the rationales.} The provision of labels shifts the model's focus from outputting correct answer to how to better explain that answer. Thus, the rationales become more fluid and richer, leading to better Comprehensiveness, Logical Coherence, Faithfulness, and Diversity. While adding label seems simple, \textbf{the quality improvement is highly dependent on the correctness of the label}. In Table \ref{prompt_rat_rationale_labelwrong}, we present the rationale generated under the wrong label. It can be observed that to accommodate this incorrect label, the generated rationale are overly redundant, chaotic, and contain semantic understanding errors, resulting in low Conciseness, Coherence, and Faithfulness. It emphasizes the importance of ensuring label accuracy during our data construction process.


Moreover, nearly all previous works using LLM to generate rationales have adopted a few-shot approach, using manually written exemplars to guide the model \cite{DBLP:conf/acl/MagisterMAMS23,DBLP:conf/emnlp/KimJKJYSS23}. In Table \ref{prompt_rat_rationale_labelexem}, we show the prompt with both a label and eight high-quality exemplars and the generated rationale. \textbf{The addition of exemplars, compared to providing only a label, does not enhance overall quality.} The mimicking and learning from examples increase the Comprehensiveness of the rationales, but also make them lengthy and rigid, causing a significant drop in Conciseness and Diversity. The damage to the diversity of rationales by exemplars is overlooked in previous works, but is crucial for LLMs' learning.

\begin{figure}[t]
\centering
\includegraphics[scale = 0.3]{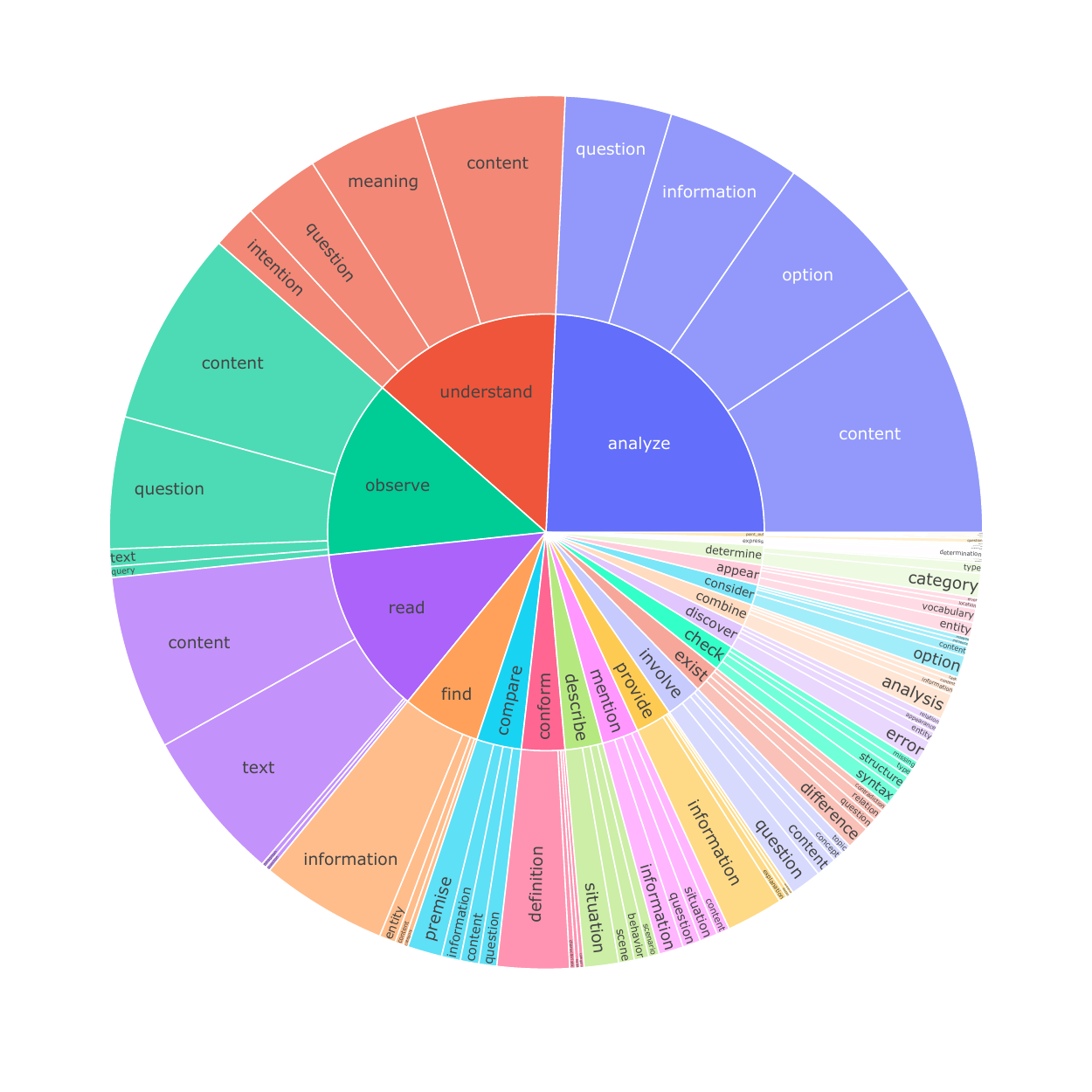}
\caption{\textbf{English} version of the top 20 common root verbs and their top 4 noun objects within the rationales.} 
\label{verb_noun_en}
\end{figure}

\begin{figure}[t]
\centering
\includegraphics[scale = 0.3]{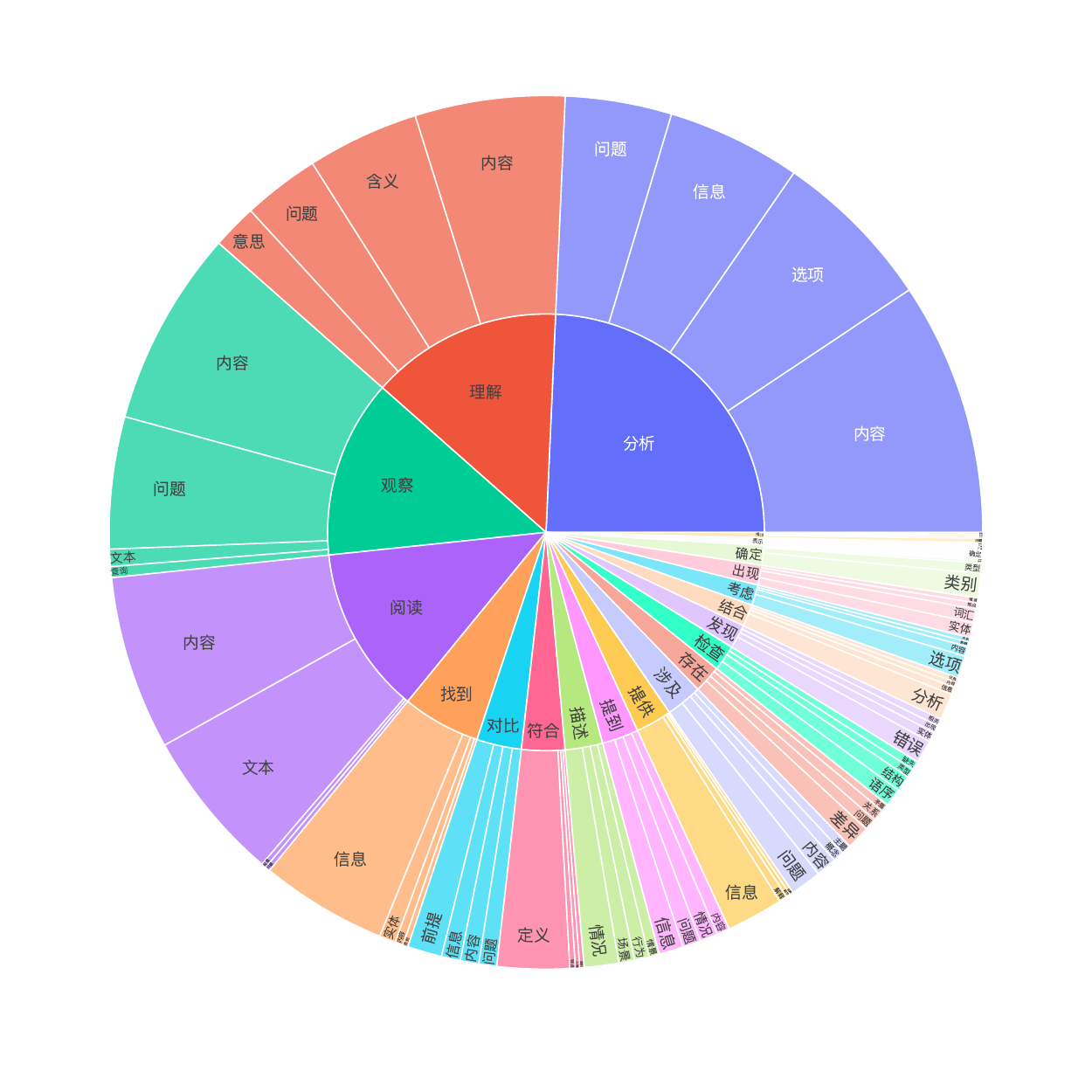}
\caption{The top 20 common root verbs (inner circle) and their top 4 noun objects (outer circle) within the rationales of our rationale collection.} 
\label{verb_noun_zh}
\end{figure}

Lastly, we explore the impact of adding judge criteria for labels. The limited label space is the most distinctive feature of NLU tasks compared to others. For every task involving categorical classification, we manually write complete and precise criteria for each label. Table \ref{prompt_rat_rationale_labelcriteria} displays the prompt and corresponding rationales. After adding criteria, \textbf{the rationales are analyzed point by point according to the given criteria, greatly enhancing Comprehensiveness and Faithfulness}. However, this also makes explanations lengthy, with similar styles and expressions across different samples, significantly reducing Conciseness and Diversity.

In summary, the inclusion of ground truth labels is crucial for the quality of rationales, hence we include them in all prompts.
Additionally, compared to including few-shot exemplars, incorporating label criteria results in a smaller decrease in Conciseness and Diversity while bringing greater improvements in Comprehensiveness and Faithfulness. Therefore, we include label criteria in 20\% of the samples to ensure high quality of both inter and intra rationales.


\begin{table*}[t]
\centering
\small
\begin{tabular}{p{15cm}}  
\hline
(Original Prompt)\newline
\textbf{Instruction}: You are an expert in question-pair matching tasks. This task originates from the Baidu ZhiDao domain, aiming to understand and compare whether the semantic content of two Chinese questions is consistent. The input includes two questions (Question 1 and Question 2), and you need to determine whether these two questions are semantically similar or express the same query intention. If the two questions fundamentally ask for the same information, their semantic relationship is matched; otherwise, it is unmatched. \newline \newline
Now, read the Question 1, Question 2 \newline 
\textbf{Input}: \newline Question 1: What's the meaning of lily? \newline Question 2: What dose lily mean?  \newline \newline
Then, think step by step, provide the reasoning process first, and finally give the relationship between them as the answer.\newline \newline

\textbf{Rationale}: \footnotesize \textcolor{red}{1. Firstly,} observing the textual expressions of Question 1 and Question 2, the main difference lies in the placement of the words "is" and "what".  Semantically, both "What's the meaning of lily?" and "What does lily mean?" are inquiries about the meaning of the word "lily". Although there is a slight difference in word order, their core intentions are consistent, as both are seeking an explanation of the word "lily". \newline  Therefore, the answer is: Matched. \\ \hline
\end{tabular}
\caption{Cases of \textbf{Original} prompt, and the generated rationale. \textcolor{red}{Red} indicates its poor Comprehensiveness and Logical Coherence.}
\label{prompt_rat_rationale_original}
\end{table*}

\begin{table*}[t]
\centering
\small
\begin{tabular}{p{15cm}}  
\hline
(Prompt with \textbf{Label})  \newline
 \textbf{Instruction} \newline 
Now, read the Question 1, Question 2, and their relationship \newline 
\textbf{Input} \newline
\textbf{Relationship}: Matched \newline 
Then, think step by step, provide the reasoning process that leads to the relationship.\newline\newline

\textbf{Rationale}: \footnotesize 1. First, by observing Question 1 "What's the meaning of lily?" and Question 2 "What does lily mean?", it is evident that the textual content of these two questions is nearly identical, with only a slight rearrangement of the words "is" and "what". Such minor differences do not affect the core semantics of the questions. \newline 2. Secondly, both questions are asking for the definition or explanation of "lily", indicating a similar intent to understand the meaning of the word "lily". \newline 3. Considering these factors, it can be determined that Questions 1 and 2 are semantically similar and express the same query intent. \newline  Therefore, the answer is: Matched. \\ \hline 
\end{tabular}
\caption{Cases of prompt with \textbf{Label}, and the generated rationale.}
\label{prompt_rat_rationale_label}
\end{table*}

\subsection{Rationale Filtering}\label{Appendix_filter_rationale}
After generating the rationales, we first apply the following rules sequentially to for filtering to ensure the quality of the rationales:
\begin{itemize}
    \item Rationales that involve safety issues and are flagged as unanswerable.
    \item Rationales that are overly long. We require that the total token count of the rationale and the label must be less than 1024.
    \item Rationales whose final answer is inconsistent with the original label. In the generation prompt, we ask the model to provide an answer judgment in the final sentence. This not only ensures the completeness of the rationales but also facilitates cleaning. If the final judgment is inconsistent with the original label, it indicates that there are likely issues in the intermediate reasoning process.
\end{itemize}
Additionally, since our rationales are ultimately used to construct instruction tuning data, we want to avoid cases where the model appears to have prior knowledge of the label. Specifically, during the reasoning process, we aim to eliminate phrases such as "supports the given label," "aligns with the given label," or "the provided label is reasonable," as they imply that the label has already been predetermined. To address this, after cleaning problematic rationales, we manually rewrite those that infer the reasoning backward from the label. We first identify rationales containing specific keywords, and then ask human annotators to either remove the corresponding sentences or rewrite the logic. Out of a dataset of 10,000 samples, approximately 50 cases contain such rationales.


\subsection{Diversity of Rationales}
To demonstrate the diversity of rationales in our rationale collection, we conduct an analysis similar in \cite{DBLP:conf/emnlp/KimJKJYSS23}. Using LTP \cite{Che_N-LTP_An_Open-source_2020}, we identify the verb closest to the root of the parse tree and extract the associated noun object. Figure \ref{verb_noun_en} and \ref{verb_noun_zh} illustrates the most frequently used verbs and their corresponding nouns. It can be observed that our rationales not only feature a variety of verbs but also associate each verb with diverse nouns, highlighting the diversity of our dataset.

\subsection{Choice of Hunyuan-Turbo} \label{Appendix_choice_hunyuan}
To validate our choice of Hunyuan-Turbo for rationale generation, we provide a manual analysis in Table \ref{tab:rationale_comparison}, comparing the rationale quality produced by different LLMs using the same prompt (with Label). The results show that the rationale quality across these top-performing closed-source models is comparable, confirming that Hunyuan-Turbo can generate high-quality rationales. Furthermore, this demonstrates that our dataset and experimental results are reproducible with rationales generated by other LLMs.

\begin{table}[t]
\centering
\begin{tabular}{lccc}
\hline
\textbf{Dimension}      & \textbf{Hunyuan} & \textbf{Qwen} & \textbf{GPT} \\ \hline
\textbf{Concise}        & 3.6                    & 3.8                 & 3.3             \\
\textbf{Comprehensive}  & 3.9                    & 3.5                 & 3.9             \\
\textbf{Cohere}         & 4.0                    & 3.9                 & 4.3             \\
\textbf{Faithful}       & 4.1                    & 4.2                 & 4.1             \\
\textbf{Diverse}        & 4.5                    & 4.1                 & 4.7             \\ \hline
\textbf{AVG}            & 4.0                    & 3.9                 & 4.1             \\ \hline
\end{tabular}
\caption{Manual analysis of rationale quality across different dimensions for Hunyuan-Turbo, Qwen-Turbo, and GPT-4o.}
\label{tab:rationale_comparison}
\end{table}

\begin{table*}[t]
\centering
\small
\begin{tabular}{p{15cm}} 
\hline
(Prompt with \textcolor{red}{Wrong} \textbf{Label})  \newline
 \textbf{Instruction} \newline 
Now, read the Question 1, Question 2, and their relationship \newline 
\textbf{Input} \newline
\textbf{Relationship}: \textcolor{red}{Unmatched} \newline 
Then, think step by step, provide the reasoning process that leads to the relationship.\newline \newline 

\textbf{Rationale}: \footnotesize 1. First, we analyze Question 1, "What's the meaning of lily?", which is asking for the meaning or definition of the word "lily". Next, we analyze Question 2, "What does lily mean?", which similarly is asking for the meaning or definition of the word "lily". \textcolor{red}{At first glance, both questions seem to be asking about the same content. However,} to ensure semantic accuracy, we need to consider two aspects: one is the expression of the question, and the other is the intent behind the question.\newline
2. In terms of expression, Question 1 uses "What's the meaning of", while Question 2 simplifies this to "What does mean". Although the expressions differ slightly, their intents are the same, i.e., to inquire about the meaning of a word. However, semantic matching depends not only on the superficial expression of the question but also on the specific context and environment in which the question is posed. \textcolor{red}{Even though the questions appear to ask the same meaning on the surface, in some situations, such as in a broader context, a more precise expression might be needed to distinguish different query intents.} \newline
3. Therefore, in this case, if we consider the subtle differences in the way the questions are expressed, and the possibility that a more explicit expression might be required in certain specific contexts, we can conclude that although both questions inquire about the same information, the differences in their expressions lead to them not being completely semantically matched.\newline
Thus, the answer is: Non-Matched.
\\ \hline 
\end{tabular}
\caption{Cases of prompt with \textcolor{red}{Wrong} \textbf{Label}, and the generated rationale. \textcolor{red}{Red} in Rationale indicates its poor Conciseness, Logical Coherence and Faithfulness. }
\label{prompt_rat_rationale_labelwrong}
\end{table*}

\begin{table*}[t]
\centering
\small
\begin{tabular}{p{15cm}}  
\hline
(Prompt with \textbf{Label} \& \textbf{8 Exemplars}) \newline
\textbf{Instruction} \newline \newline 
\textbf{[Exemplar 1]} \newline \footnotesize
Question 1: Looking for e-books of novels. Question 2: Looking for the e-book of the novel "Empresses in the Palace"!  Relationship: Unmatched 
\newline Rationale: 1. Analyze the key information in Question 1: The key information in Question 1 is "e-books of novels," which is a very broad request without specifying any particular novel name or genre. Thus, it can be understood as asking for an e-book version of any novel.
2. Analyze the key information in Question 2: The key information in Question 2 is "e-book of the novel 'Empresses in the Palace'," which is a specific request specifying the novel name as "Empresses in the Palace." Hence, it is asking for the e-book version of this specific novel.
3. Compare the key information of both questions: The objects of the requests in Question 1 and Question 2 are different. Question 1 is asking for an e-book of any novel, while Question 2 is asking for an e-book of the specific novel "Empresses in the Palace."
4. Determining if the semantic content is consistent: Although both questions are about asking for e-book versions of novels, Question 1 does not specify a particular novel, whereas Question 2 specifies the novel "Empresses in the Palace." Therefore, the semantic content of the two questions is not consistent.
5. Conclusive analysis: Based on the above analysis, Questions 1 and 2 differ fundamentally in the information being queried. Question 1 is a general request, while Question 2 is a specific request. The intent behind the queries is different. Therefore, the answer is: Unmatched.\newline
\textbf{[Exemplar 2]}... \newline 
\newline
Now, read the Question 1, Question 2, and their relationship \newline
\textbf{Input} \newline
\textbf{Relationship}: Matched \newline 
Then, refer to the above exemplars, provide the reasoning process that leads to the relationship.\newline\newline

\textbf{Rationale}: \footnotesize \textcolor{red}{1. } Analysis of Key Words and Intent in Question 1: The key words in Question 1 are "lily" and "what's the meaning," with the intent being to inquire about the meaning or explanation of the word "lily."\newline 
\textcolor{red}{2. } Analysis of Key Words and Intent in Question 2: The key words in Question 2 are also "lily" and "what does mean," with the intent similarly being to inquire about the meaning or explanation of the word "lily."\newline 
\textcolor{red}{3. } Comparison of Semantic Content in Both Questions: Although Questions 1 and 2 use different phrases \textcolor{red}{("what's the meaning" vs. "what does mean")}, they both focus on asking about the meaning of the word "lily."\newline 
\textcolor{red}{4. } Assessment of Whether Both Questions Express the Same Query Intent: Since both questions seek an explanation or meaning of the word "lily," \textcolor{red}{even though expressed slightly differently,} their query intents are consistent.\newline 
\textcolor{red}{5. } Overall Judgment on the Match Relationship Between the Two Questions: Combining all the information above, Questions 1 and 2 fundamentally inquire about the same thing, which is the meaning of the word "lily." \newline
Thus, the conclusion is: Matched.
\\ \hline
\end{tabular}
\caption{Cases of prompt with \textbf{Label} and eight \textbf{Exemplars}, and the generated rationale. \textcolor{red}{Red} in Rationale indicates its poor Conciseness. }
\label{prompt_rat_rationale_labelexem}
\end{table*}

\begin{table*}[t]
\centering
\small
\begin{tabular}{p{15cm}}  
\hline
(Prompt with \textbf{Label} \& \textbf{Label Criteria})  \newline
\textbf{Instruction} \newline \newline
\textbf{Label Criteria:} \newline \footnotesize
Matched: A relationship between two questions is considered matched if any of the following conditions are met:
Similar Intent: Both questions explicitly express the same query intent, even if their literal expressions or detailed descriptions differ.
Same Core Query: Despite possible differences in peripheral information, both questions have the same core subject or objective.
Interchangeability: The answer to one question can reasonably serve as the answer to the other, meaning these two questions can be interchangeably used in practical scenarios. 
\newline
Unmatched: A relationship between two questions is considered not a match if any of the following conditions are met:
Different Intent: The two questions express different query intents, asking fundamentally different information or matters.
Different Core Query: Even if there are similarities in some aspects, the core focus or main query target of the two questions differs.
Answers are Not Interchangeable: The answer to one question cannot reasonably serve as the answer to the other, indicating a fundamental difference in intent or requirement between the two questions. \newline
Now, read the Question 1, Question 2, and their relationship \newline 
\textbf{Input} \newline
\textbf{Relationship}: Matched \newline 
Then, step by step, provide the reasoning process that leads to the relationship.\newline\newline

\textbf{Rationale}: \footnotesize \textcolor{red}{1. } First, let's analyze Question 1: "What's the meaning of lily?" and Question 2: "What does lily mean?". Literally, these two questions are very similar, with only the positions of "is" and "what" switched. This difference does not affect the core query intent.\newline
\textcolor{red}{2. }Secondly, from the perspective of query intent, both questions are asking about the meaning of the word "lily". Whether it is \textcolor{red}{"What's the meaning of lily?" or "What does lily mean?"}, both are seeking an explanation or definition of the word "lily".
\newline
\textcolor{red}{3. }Furthermore, considering the interchangeability of the answers, if one question is answered, the answer to the other question is the same. For example, if the answer to "What's the meaning of lily?" is "A lily is a type of plant", then the answer to "What does lily mean?" would also be "A lily is a type of plant".\newline
Therefore, combining the above analysis, the semantic relationship between these two questions is matched. Thus, the answer is: Matched.
\\ \hline
\end{tabular}
\caption{Cases of prompt with \textbf{Label} and eight \textbf{Label Criteria}, and the generated rationale. \textcolor{red}{Red} in Rationale indicates its poor Conciseness.}
\label{prompt_rat_rationale_labelcriteria}
\end{table*}

\begin{table*}[t]
\centering
\footnotesize
\begin{tabular}{p{1.5cm} p{13cm}}
\hline
\textbf{Dimensions}           & \textbf{Scoring Criteria}                                                                                                                                                                                                                                                                                                                                                                                                                                                                                                                                                                                                   \\ \hline
Conciseness       & \begin{tabular}[c]{@{}p{13cm}@{}} 
\textbf{Score 5}: Rationales are highly concise, with no unnecessary details or repetition; every word adds value.\\ 
\textbf{Score 4}: Rationales are mostly concise but contain minor redundancy or slightly excessive detail.\\ 
\textbf{Score 3}: Rationales are moderately concise; include some unnecessary sentences or repetition that could be trimmed.\\ 
\textbf{Score 2}: Rationales are verbose, with significant redundancy or irrelevant details that detract from clarity.\\ 
\textbf{Score 1}: Rationales are excessively lengthy or repetitive, obscuring the key points entirely.
\end{tabular} \\ \hline
Comprehen\newline siveness & \begin{tabular}[c]{@{}p{13cm}@{}} 
\textbf{Score 5}: Rationales fully cover all essential aspects and relevant background information; nothing critical is omitted.\\ 
\textbf{Score 4}: Rationales cover most key aspects but have minor omissions or areas lacking sufficient detail.\\ 
\textbf{Score 3}: Rationales cover some key aspects but miss important elements or lack context in key areas.\\ 
\textbf{Score 2}: Rationales are incomplete, omitting multiple critical points or background information necessary for clarity.\\ 
\textbf{Score 1}: Rationales fail to address the core question; critical omissions make the rationale ineffective.
\end{tabular} \\ \hline
Logical \newline Coherence         & \begin{tabular}[c]{@{}p{13cm}@{}} 
\textbf{Score 5}: Rationales follow a fully coherent, step-by-step logical structure, with no contradictions or leaps in reasoning.\\ 
\textbf{Score 4}: Rationales are mostly logical, with minor lapses in coherence or occasional steps that feel abrupt or disconnected.\\ 
\textbf{Score 3}: Rationales have noticeable gaps or inconsistencies in the reasoning process but maintain overall direction.\\ 
\textbf{Score 2}: Rationales are poorly structured, with frequent logical gaps, contradictions, or unclear progression.\\ 
\textbf{Score 1}: Rationales lack logical coherence entirely, with random or disconnected reasoning.
\end{tabular} \\ \hline
Faithfulness      & \begin{tabular}[c]{@{}p{13cm}@{}} 
\textbf{Score 5}: Rationales are entirely faithful to the input and label, fully grounded in the evidence without deviation.\\ 
\textbf{Score 4}: Rationales are mostly faithful but include minor inaccuracies or irrelevant points that don't detract significantly.\\ 
\textbf{Score 3}: Rationales have moderate faithfulness; partially align with the input and label but include notable inaccuracies.\\ 
\textbf{Score 2}: Rationales stray significantly from the input or label, with frequent inaccuracies or irrelevant reasoning.\\ 
\textbf{Score 1}: Rationales are unfaithful to the input or label, with justification that is incorrect or fabricated.
\end{tabular} \\ \hline
Diversity         & \begin{tabular}[c]{@{}p{13cm}@{}} 
\textbf{Score 5}: Rationales are highly diverse, showcasing varied reasoning styles, phrasing, or perspectives.\\ 
\textbf{Score 4}: Rationales are mostly diverse but include some recurring patterns or similar phrasing across examples.\\ 
\textbf{Score 3}: Rationales show moderate diversity; some variety is present, but significant repetition is noticeable.\\ 
\textbf{Score 2}: Rationales lack diversity, with repeated reasoning styles or identical phrasing dominating the responses.\\ 
\textbf{Score 1}: Rationales are entirely repetitive, offering no variation in reasoning, style, or phrasing. 
\end{tabular}   \\ \hline                                        
\end{tabular}
\caption{Human evaluation scoring criteria for assessing the quality of rationales}
\label{app:scoring_criteria}
\end{table*}

\clearpage

\clearpage

\begin{figure*}[t]
\centering
\includegraphics[scale = 0.3]{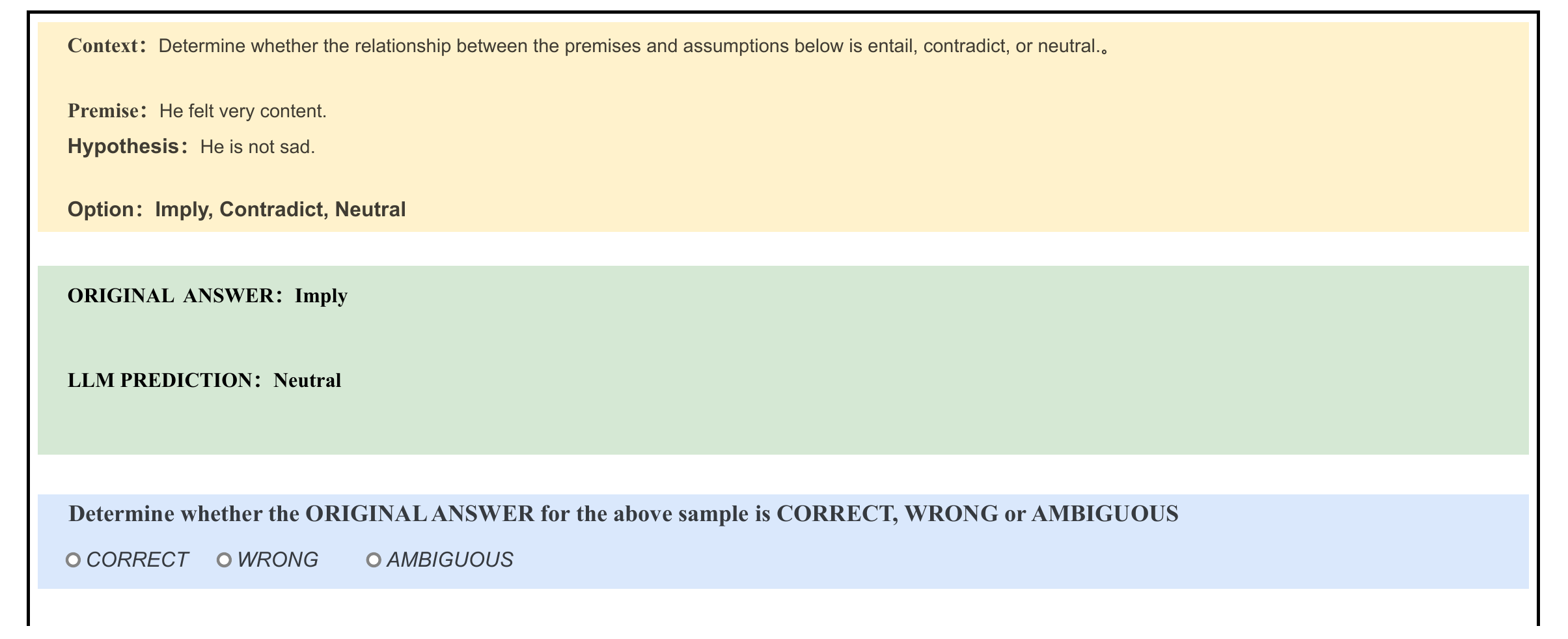}
\caption{Screenshot of user interface for \textbf{judging label accuracy}, showing the sample input with original label and the LLMs' prediction, which are inconsistent.} 
\label{screenshot_cor}
\end{figure*}

\begin{figure*}[t]
\centering
\includegraphics[scale = 0.3]{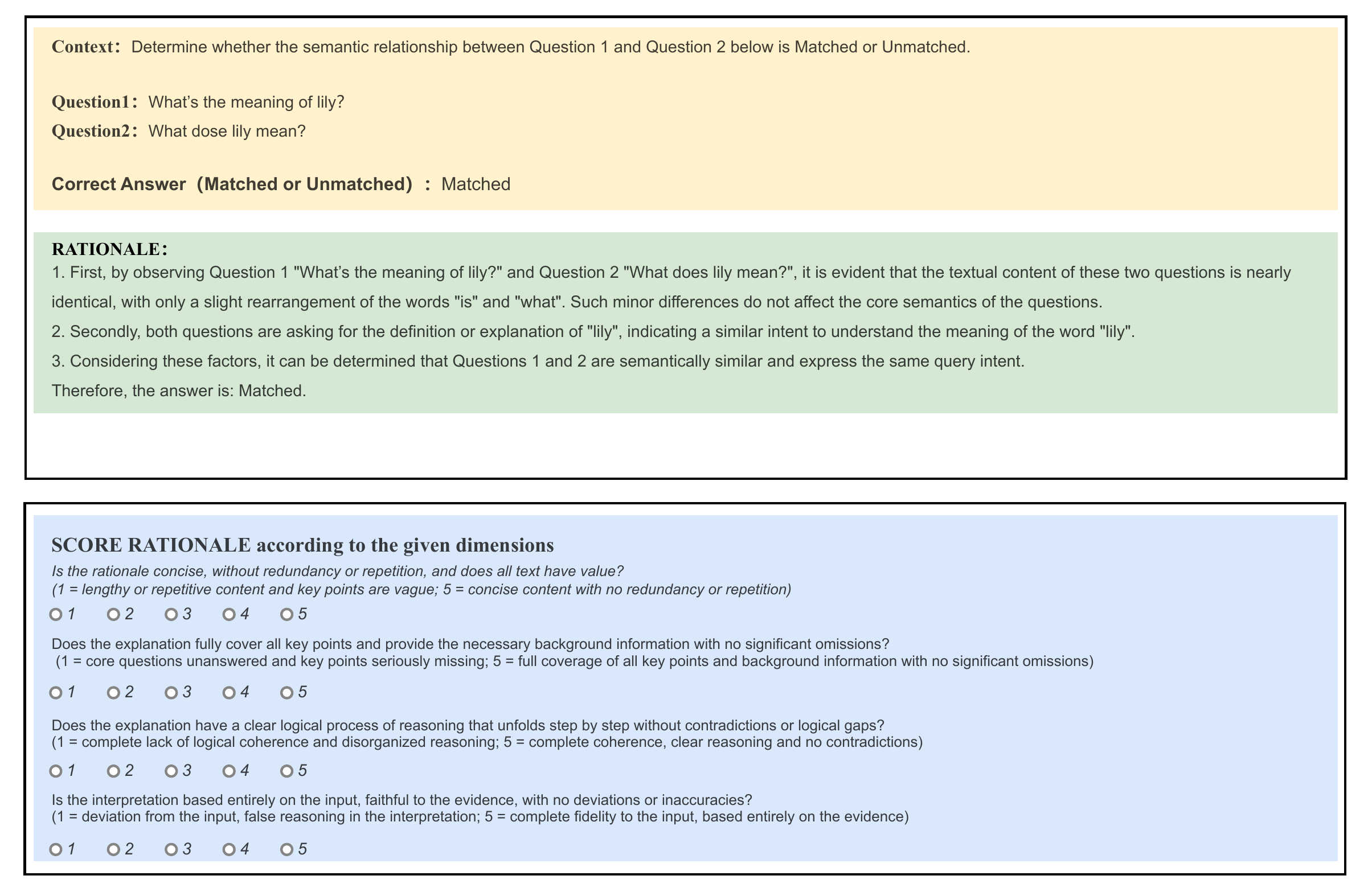}
\caption{Screenshot of user interface for  \textbf{scoring rationale quality}, showing the sample input with label, the generated rationale and the scoring dimension.} 
\label{screenshot_dimension}
\end{figure*}

\begin{figure*}[t]
\centering
\includegraphics[scale = 0.3]{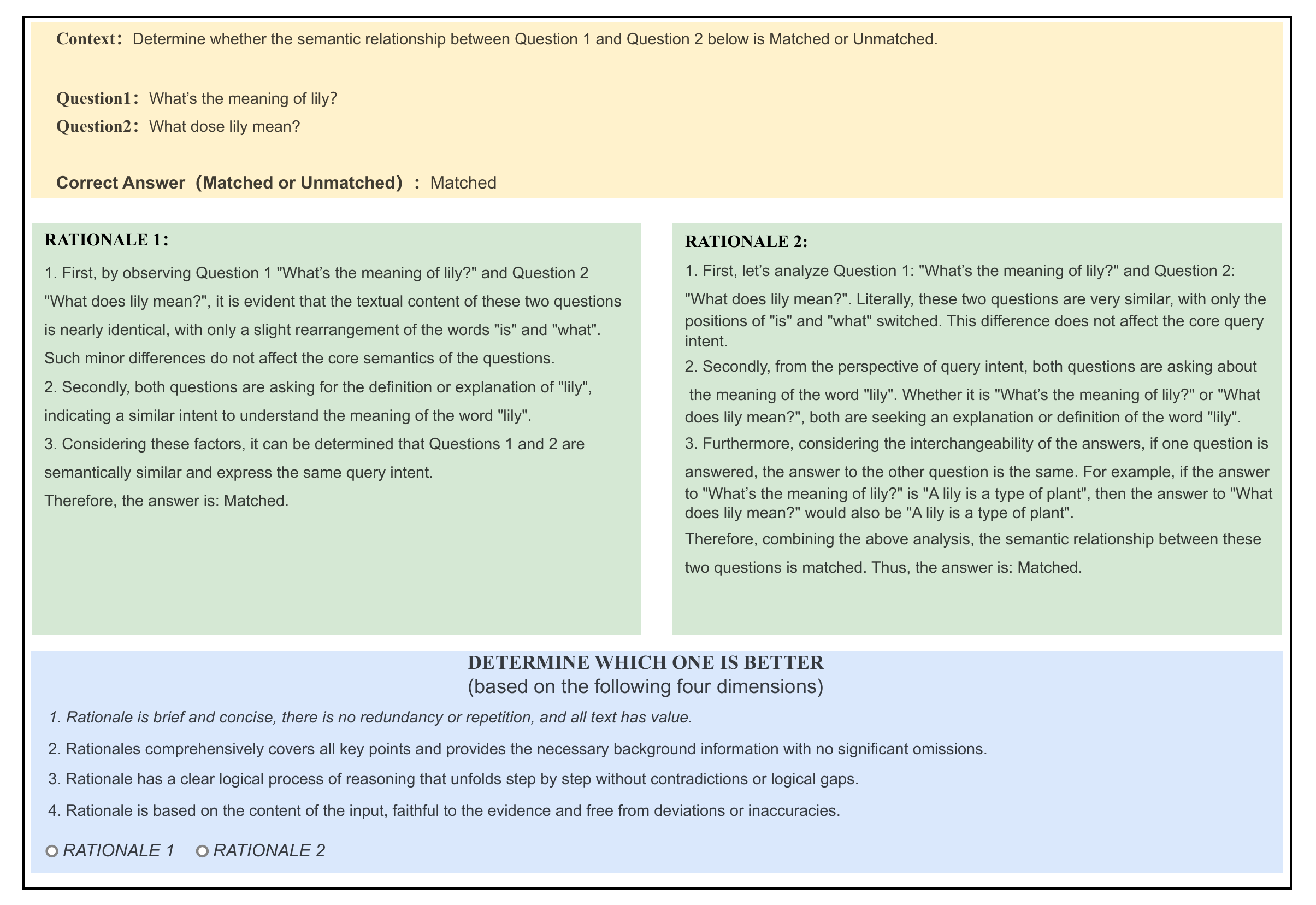}
\caption{Screenshot of user interface for \textbf{judging  rationale comparison}, showcase the sample input and label, with two rationales for comparison.} 
\label{screenshot_judge_qua}
\end{figure*}

\section{Human Annotator}\label{Appendix_human_annotator}
In this work, we have human annotators determining label accuracy, scoring rationale quality, categorizing CoT errors and comparing rationale all require domain expertise. These tasks all require domain expertise and anonymous crowd workers are unable to perform them effectively. Therefore, we ask collect-graduate Chinese-speaking students with extensive knowledge in NLP as expert annotators, to complete the annotations.

Figure \ref{screenshot_cor} illustrates the UI used to annotate whether the original labels of the samples are correct, incorrect, or ambiguous (\S \ref{NLU_Clean} and Appendix \ref{details_in_human_analysis}). Figure \ref{screenshot_dimension} shows the UI used to assign scores to different quality dimensions (Appendix \ref{Appendix_prompt_design_experiment}). Each sample is scored on four dimensions, while the diversity dimension is evaluated after assessing samples within the same dataset. Figure \ref{screenshot_judge_qua} illustrates the UI used to compare rationales generated by different models (\S \ref{experiment_tuning}).

\section{Experiments }\label{Appendix_seen}
\subsection{Training and Evaluation Details}\label{Appendix_Experimental_Setups}
We conduct our experiments using 8 NVIDIA A100 GPUs (40GB) with bf16 precision for training and evaluation. 
For experiments for RQ1 and RQ2, we train the model in a task and then evaluate on the same seen task, reporting the average results across 3 runs.Given a task, we train on all the datasets included in the task, and then evaluate on the test set of each dataset then take the macro average as the performance.
And for experiments for RQ3, we instruction-tune the model in the full NLURC data except for the evaluation task, reporting the results of 1 run.
Since LLMs tend to favor certain options \cite{DBLP:conf/iclr/Zheng0M0H24}, we retain the original labels and use specific answer extraction parsers for each dataset.
The models are trained using LoRA \cite{DBLP:conf/iclr/HuSWALWWC22}, targeting all matrix in layers and modules while keeping other parameters at the default settings provided by HuggingFace. The learning rate is set to 5e-5, with a warm-up ratio of 0.02 and a cosine learning rate scheduler. The batch size and gradient accumulation steps are adjusted according to the model size. Both the maximum input and output lengths are set to 1024 tokens. Each model is trained for 5 epochs, with the best-performing checkpoint on the development set used for evaluation on the test set. Training is implemented using the LLaMA-Factory \cite{zheng2024llamafactory}, and inference is performed using vLLM \cite{kwon2023efficient}.

For the inference settings, we use a temperature of 0 for Direct inference. For CoT inference, we employ both a temperature of 0 and a temperature of 0.7 with top\_p=0.9, reporting the results from the configuration with better performance. Regarding metrics, Accuracy is adopted for all tasks except for sequence labeling tasks, such as MRC and NER, where the F1 metric is used instead.

In Figure \ref{method}, we illustrate the organization of training samples using different training methods. And in Table \ref{Prompt_methods}, we present instruction prompts under different methods, using the RC. w/CS. task as an example. Notably, for the Reason prompt, the input additionally includes "Let's think step by step" in the end to explicitly guide the model to generate reasoning first. Training samples for the Mix and Align methods incorporate instruction prompts from both the Label-Only and Reason.

\subsection{More Results}
We present the performances of the 0.5B, 7B, and 32B models on five tasks using different methods in Table \ref{results_all}. In Table \ref{app:results_all_sup}, we further show the results for the 1.8B models. It can be seen that Align consistently outperforms Label-Only, which in turn outperforms others, confirming that our previous conclusion holds.

\begin{table*}[t]
\centering
\small
\begin{tabular}{p{1cm}p{13.5cm}}  
\hline
Method & \textbf{Instruction Prompt}: \newline Input: Question 1: ; Question 2: ;\newline Determine the Relationship between the following Question 1 and Question 2. \\ \hline

Label-Only & (Input: ...; Determine ...;) Directly output Matched or Unmatched as the answer. \\ \hline
\textbf{Reason}& (Input: ...; Determine ...;) Give the reasoning process first, and ends with “Therefore, the answer is :” to provide Matched or Unmatched as the answer. \\ \hline
\textbf{Explain} & (Input: ...; Determine ...;) Directly output Matched or Unmatched as the answer, and then give the reasoning process. \\
\hline
\end{tabular}
\caption{Instruction Prompt for a Paraphrase dataset under different methods. }
\label{Prompt_methods}
\end{table*}

\subsection{CoT Error Types and Analysis}\label{Appendix_wrong_reasoning}
To investigate why providing the reasoning process before generating answers is ineffective or even harmful in NLU tasks, we conduct a quantitative experiment and a case study. 
We select samples from a 7B-sized model trained with Align, specifically those that are correct in Direct inference but incorrect in CoT inference.
We randomly select samples from Stance Detection and Reading Comprehension with Common Sense task, and ask human annotators to categorize the CoT errors to one of the four types. The error type definitions are in Table \ref{Error_type_def}.
And in Table \ref{Reason_Reason_Wrong_Case}, we provide cases of understanding errors in the two tasks.
We find that the reasoning processes are complete and logically consistent, but there are semantic understanding errors. These minor but crucial errors lead to subsequent reasoning mistakes, resulting in incorrect final answers.

\begin{table*}[t]
\centering
\begin{tabular}{p{2.5cm}p{12.5cm}}
\hline
\textbf{Error Type}         & \textbf{Definition}                                                                                                                                               \\ \hline
\textbf{Understanding}                & Understanding errors occur when the model fails to correctly analyze the meaning of input due to missing or misapplying commonsense knowledge, factual information, or implicit meaning (e.g., sarcasm, cultural references, social norms, physical realities). The model may produce an answer that contradicts basic real-world principles or established facts.
\\ \hline
\textbf{Logical} & Logical errors occur when the model’s reasoning process is flawed despite correctly understanding the input and context. Mistakes include faulty deductions, incorrect generalizations, invalid cause-and-effect relationships, or inconsistencies in multi-step reasoning. The model misapplies logical principles rather than misinterpreting language.                              \\ \hline

\textbf{Context} & Context errors occur when the model incorrectly uses or ignores relevant contextual information. Errors can stem from failing to connect related sentences, mislocating the key part of input, misattributing references, or overlooking prior discourse. The model may focus on irrelevant details while neglecting crucial background or supporting evidence.                         \\ \hline
\textbf{Linguistic}      & Linguistic errors occur when the model fails to correctly analyze the grammatical or structural aspects of language. These errors include incorrect sentence parsing (e.g., misidentifying clause boundaries, negations, or syntactic dependencies), failure to resolve pronoun references, improper handling of fixed expressions, and errors in morphological agreement (e.g., incorrect tense or number in morphologically rich languages).                \\ \hline
\end{tabular}
\caption{Definitions of Error Types}
\label{Error_type_def}
\end{table*}

\begin{table*}[]
\centering
\small
\begin{tabular}{p{1.6cm}p{5.2cm}p{7.8cm}}  
\hline
Task & Instruction \& Text                                                                                                                                                                                                                                                                                                                                                                               & Reasoning Process \& Answer                                                                                                                                                                                                                                                                                                                                                                                                                                                                        \\ \hline

Stance \newline NLPCC  & \textbf{Instruction}: Determine whether the Text takes a neutral, supportive, or against stance on the given Comment Object. \newline \textbf{Text}: Support, was hit by an electric cycle on the sidewalk. \newline \textbf{Comment Object}: Ban on moto- and electric-cycle.\newline \textbf{Answer}: supportive & \footnotesize 1. First, we analyze the content of the text. The text mentions “support”, which indicates that \textcolor{red}{the author supports the behavior of electric cycles driving on the sidewalk}. 2. Next, we need to relate this position to the subject of the comment. The ban is intended to restrict the movement of electric cycles in the city to protect the safety of pedestrians and vehicles. The support for electric cycles on sidewalks mentioned in the text is contrary to the goal of the ban on electric cycles. 3. Since the supportive stance mentioned in the text is contrary to the objective of the comment, we can infer that the author takes an opposing stance to the comment object. \newline Therefore, the answer is: \textcolor{red}{against}.
\\
\hline
RC. w/CS. \newline C3  & \textbf{Instruction}: Read the Background and Question below, choose the most appropriate answer from the Options. \newline
\textbf{Background}: Woman: What time does your company usually finish work? Man: It’s 4:30 on Thursdays, and 5:00 on other days. If we work overtime on weekends, I'm not sure.\newline \textbf{Question}: What time does the man finish work on Saturdays? \newline \textbf{Options}: A. 5:00; B. 5:30; C. 4:30; D. Uncertain. \newline \textbf{Answer}: D. Uncertain & \footnotesize 1.First, we extract key information from the conversation. The man mentions, "Thursday is 4:30, other days are 5:00," indicating that, except for Thursday, the usual workday ends at 5:00. He then adds, "If there's overtime on the weekend, I'm not sure," meaning weekend overtime hours are uncertain. 2. Next, we analyze the question: "What time does the man finish work on Saturday?" \textcolor{red}{Since he didn't specify Saturday's hours but mentioned uncertainty only in the case of overtime, we can infer that without overtime, the Saturday hours are fixed}. Given that other weekdays end at 5:00, it’s reasonable to conclude that, without overtime, Saturday also ends at 5:00. \newline Therefore, the answer is: \textcolor{red}{A. 5:00}. \\
\hline
\end{tabular}
\caption{Cases from Align-trained 7B models where CoT fails but Direct inference succeeds, categorized as Understanding errors. Errors in the reasoning process are highlighted in red.}
\label{Reason_Reason_Wrong_Case}
\end{table*}

\begin{table*}[t]
\centering
\begin{tabular}{lllcccccc}
\hline
Sizes & Infer & Train & Read. Comp. & Stance & Topic & Linguistic & RC. w/CS. & Average \\ \hline
\multirow{10}{*}{1.8B} & \multirow{4}{*}{CoT}   & Original & 38.20 & 45.36 & 52.27 & 22.79 & 52.18 & 42.16 \\ 
&& Reason      & 73.69       & 77.57  & 83.91 & 52.71   & 51.88    & 67.95 \\ 
&& Mix      & 76.94       & 81.85  & 86.99 & 53.87   & 51.70     & 70.27 \\  
&& Align     & 78.37       & 83.14  & 88.01 & 55.30    & 57.18    & 72.40 \\ \cdashline{2-9}

& \multirow{6}{*}{Direct} & Original & 42.47 & 55.26 & 43.3 & 24.01 & 46.03 & 42.21 \\  
&& Label-Only      & \underline{82.25}       & \underline{89.08}  & \underline{93.41} & \underline{69.22}   & \underline{68.79}    & \underline{80.55} \\
&& Explain      & 78.18       & 85.17  & 91.30  & 62.36   & 60.20     & 75.44 \\
&& Mix      & 80.72       & 87.13  & 92.74 & 66.22   & 62.99    & 77.96 \\ 
&& \textbf{Align}     & \textbf{84.11}       & \textbf{90.43}  & \textbf{93.70}  & \textbf{71.80}    & \textbf{77.41}    & \textbf{83.49} \\ \hline


\end{tabular}
\caption{More results of Table \ref{results_all}. Results of different combinations of training and inference methods for model size of 1.8B. }
\label{app:results_all_sup}
\end{table*}

\begin{table*}[t]
\centering
\begin{tabular}{llcccccc}
\hline
\multicolumn{2}{c}{Sizes/Datasets} & Read. Comp. & Stance & Topic & Linguistic & RC. w/CS. & Average \\ \hline
\multirow{6}{*}{0.5B}     & 0 (Reason)     & 40.54       & 44.03  & 27.73 & 10.15   & 5.56     & 25.60  \\
                          & 0.25   & 79.01       & \textbf{86.20}  & 91.99 & 63.56   & 57.25    & 75.60  \\
                          & 0.5     & 79.47       & 86.11  & \textbf{92.08} & \textbf{64.69}   & \textbf{63.48}    & \textbf{77.17}  \\
                          & 0.75   & \textbf{79.69}       & 85.60  & 91.80 & 63.15   & 63.29    & 76.71  \\
                          & 1 (Label-Only)     & 77.80       & 83.20  & 91.55 & 57.77   & 56.61    & 73.39  \\ \hline
\multirow{6}{*}{1.8B}     & 0 (Reason)      & 58.52       & 65.03  & 75.63 & 34.99   & 16.59    & 50.15  \\
                          & 0.25   & 84.16       & 89.83  & 93.53 & 70.87   & 69.86    & 81.65  \\
                          & 0.5    & \textbf{84.21}       & \textbf{91.23}  & \textbf{93.70} & \textbf{71.80}   & \textbf{77.41}    & \textbf{83.67}  \\
                          & 0.75   & 83.69       & 91.08  & 92.86 & 71.29   & 76.42    & 83.07  \\
                          & 1 (Label-Only)     & 82.25       & 89.08  & 93.41 & 69.22   & 68.79    & 80.55  \\ \hline
\multirow{6}{*}{7B}       & 0 (Reason)      & 73.83       & 80.27  & 90.74 & 53.05   & 22.41    & 64.06  \\
                          & 0.25   & 87.68       & 93.66  & 94.57 & 76.30   & 88.22    & 88.09  \\
                          & 0.5    & \textbf{87.72}       & 94.05  & \textbf{94.59} & \textbf{76.72}   & \textbf{90.57}    & \textbf{88.72}  \\
                          & 0.75   & 87.50       & 94.30  & 94.50 & 76.52   & 90.24    & 88.60  \\
                          & 1 (Label-Only)     & 87.46       & \textbf{94.41}  & 94.54 & 76.68   & 89.45    & 88.51  \\ \hline
\end{tabular}
\caption{Effects of label learning in NLU tasks. Performances of varying coefficients for labels and rationales in Align training . The value indicates the coefficient for label. 0 and 1 actually denote Reason and Label-Only training, respectively. }
\label{align_abla}
\end{table*}

\subsection{Effects of Label Learning in NLU tasks}\label{Appendix_effect_label_learning}
Align training separately calculates the loss for samples containing only labels and samples containing both rationales and labels, and then weights these losses with specific coefficients (the sum of the weights equals 1). To explore the importance of labels and rationales in this learning process, we conduct ablation experiments on the loss coefficients. Table \ref{align_abla} shows the performance of Direct inference after training with label coefficients of 0, 0.25, 0.5, 0.75, and 1. It's noted that label coefficients of 0 and 1 correspond to Reason and Label-Only training, respectively. It's found that the best performances mostly occur at a coefficient of 0.5, indicating that a balanced learning of both aspects is optimal for NLU tasks. Moreover, the performances with coefficients of 0 and 0.25 are generally better than those with coefficients of 1 and 0.75, suggesting that label learning should always be at the core of NLU task training.


\begin{table*}[]
\centering
\small
\begin{tabular}{p{1cm}p{13.6cm}}  
\hline
Model & \textbf{Background:} You can find your favorite tea everywhere in Zhabei District. Although Shanghai does not produce tea, it is the largest tea consumption and distribution city in China. \newline \textbf{Question:} Select the option consistent with the content of the background. \newline \textbf{Options:} A. Most of Shanghai's tea is exported abroad; B. Tea is cultivated in Zhabei District, Shanghai; C. Shanghai has a large tea consumption; D. Shanghai is the largest tea-producing area in China. \textbf{Answer}: C. Shanghai has a large tea consumption \\ \hline
Hunyuan & \footnotesize 1. First, we carefully read the background information. It provides two key pieces of information: Shanghai has a large tea consumption, and tea distribution is highly active. \newline
2. Next, let's analyze each option one by one:
\textbf{A.} It is not supported by the background information. The background only mentions that Shanghai is a center for tea consumption and distribution, without indicating that tea is mainly exported abroad.
\textbf{B.} It contradicts the background information. The background clearly states that Shanghai does not produce tea, so it is impossible for Zhabei District to cultivate tea.
\textbf{C.} It aligns with the background information. The background explicitly points out that Shanghai is the largest tea consumption city in China, indicating that tea consumption is indeed large.
\textbf{D.} It contradicts the background information. The background clearly states that Shanghai does not produce tea. \newline
3. Based on the above analysis, we conclude that the option consistent with the question is C. \\ \hline
72B & \footnotesize According to the background, Shanghai does not produce tea, but it is the largest city in China for tea consumption and distribution. This suggests that there is a huge demand for tea within Shanghai, i.e. consumption is high. Option \textbf{A} mentions selling to foreign countries, which is not mentioned in the context; Option \textbf{B} mentions growing tea, which contradicts the fact that Shanghai does not produce tea in the context; and Option \textbf{D} incorrectly claims that Shanghai is the largest producer of tea in China, contrary to the contextual information. Therefore, based on the fact that Shanghai is the center of tea consumption and distribution in China, it can be inferred that the correct answer is C. \\ \hline
7B-Seen & \footnotesize 1. First, we need to carefully analyze the background information. It mentions that Shanghai does not produce tea, but it is the largest city in China for tea consumption and distribution. This means that the consumption of tea in Shanghai is very large, far exceeding that of other regions. \newline
2. Next, let's consider each option:
\textbf{A.} It is not clearly supported by the background information, so its correctness cannot be determined.
\textbf{B.} It contradicts the background information because the background clearly mentions that Shanghai does not produce tea.
\textbf{C.} It matches the background information “Shanghai does not produce tea, but it is the largest city in China in terms of tea consumption and distribution”.
\textbf{D.} It contradicts the background information, because the background clearly mentions that Shanghai does not produce tea.\newline
3. Summarizing the above analysis, we conclude that Shanghai consumes a lot of tea, and the answer is C. \\ \hline

7B-Unseen & \footnotesize 1. We need to understand the context of the question. The background mentions that Shanghai does not produce tea, but is the largest city in China for tea consumption and distribution. This indicates that Shanghai consumes a lot of tea.\newline
2. Next, we analyze the options:
\textbf{A.} It doesn't mention the consumption of tea, so it can't be determined.
\textbf{B.} It has nothing to do with the context of the question, so it cannot be determined.
\textbf{C.} It is consistent with the context of the question and suggests that Shanghai consumes a lot of tea.
\textbf{D.} It contradicts the context of the question because it is mentioned in the context that Shanghai does not produce tea. \newline
3. Summarizing the above analysis, we can conclude that the answer which is consistent with the context of the question is C. \\
\hline
\end{tabular}
\caption{Rationale comparisons on $C^3$ dataset (RC. w/ CS) among different models.}
\label{app:rationale_comp}
\end{table*}

\subsection{Quality Analysis of Rationales from Different Models}\label{app:unseen_rat_analyses}
In \S \ref{experiment_tuning}, we present a comparison of the quality of rationales generated by four models: Qwen-7B model trained on NLURC data excluding evaluation tasks (7B-Unseen), original Hunyuan-turbo (Hunyuan-Original), original Qwen-72B (72B-Original), and Qwen-7B trained solely on the evaluation task (7B-Seen).

It covers four distinct evaluation tasks, Reading Comprehension, Stance Detection, Chinese Linguistics, and Reading Comprehension with Common Sense. For each task, we randomly sample 50 rationales generated by different models and ask human annotators  to compare them pairwise for evaluation.
All models are prompted to provide only the rationale for the given correct answer using the Explain inference.
In this subsection, we conduct a case study, detailed in Table \ref{app:rationale_comp}, to analyze their strengths and weaknesses.
First, compared to the rationales generated by Hunyuan-Original, which consistently performs best across all dimensions, the rationales from 7B-Unseen are largely on par, except for slightly weaker logical coherence. Additionally, the rationales from the 72B-Original, despite being ten times larger than the 7B model, exhibit significantly poorer comprehensiveness and logical coherence. Lastly, the rationales generated by 7B-Seen, although stylistically similar to those of Hunyuan-Original, contain less substantive content. In contrast, the 7B-Unseen rationales show improvements in conciseness, logical coherence, and faithfulness.
This demonstrates that task and rationale diversity provide the model with richer information and perspectives, which are more effective in improving its interpretability.
In summary, our rationale collection, when used as instruction tuning data, greatly enhances the explanatory capabilities of the model.

\begin{table*}[p]
\centering
\small
\begin{tabular}{lccccc}
\hline
Dataset & |Train| & |Dev| & |Test| & Rat\_Len & Rat\_len(task) \\
\multicolumn{6}{c}{Sentiment Analysis} \\
\hdashline
ChnSentiCorp & 7,600 & 760 & 748 & 274.9 & \multirow{4}{*}{271.6} \\
SMP-EWECT & 9,993 & 1,259 & 2,412 & 232.6 & \\
ASAP-ASPECT & 9,983 & 2,144 & 2,146 & 245.8 & \\
BDCI\_Car & 4,466 & 551 & 562 & 290.1 & \\ \hline

\multicolumn{6}{c}{Stance Detection} \\
\hdashline
NLPCC-stance & 1,589 & 192 & 225 & 259.9 & \multirow{3}{*}{241.8} \\
DuReader\_yesno & 9,957 & 2,123 & 2,122 & 204.0 & \\
ReCO & 9,875 & 2,290 & 2,331 & 261.6 & \\ \hline

\multicolumn{6}{c}{Natural Language Inference} \\
\hdashline
OCNLI & 9,994 & 1,073 & 1,155 & 341.0 & \multirow{2}{*}{350.8} \\
CMNLI & 9,283 & 1,190 & 1,202 & 360.7 & \\ \hline

\multicolumn{6}{c}{Paraphrase Detection} \\
\hdashline
AFQMC & 9,982 & 1,576 & 1,525 & 291.8 & \multirow{5}{*}{325.6} \\
BUSTM & 9,968 & 2,756 & 2,775 & 278.5 & \\
LCQMC & 9,997 & 2,895 & 2,812 & 257.1 & \\
PAWS & 9,938 & 759 & 764 & 436.9 & \\
QBQTC & 9,070 & 1,391 & 1,432 & 363.7 & \\ \hline

\multicolumn{6}{c}{Coreference Resolution} \\
\hdashline
CLUEWSC2020 & 1,103 & 126 & 143 & 279.7 & 279.7 \\ \hline

\multicolumn{6}{c}{Reading Comprehension} \\
\hdashline
CMRC2018 & 9,238 & 1,117 & 1,529 & 186.2 & \multirow{3}{*}{221.7} \\
DuReader\_Robust & 9,925 & 327 & 609 & 205.4 & \\
DuReader\_checklist & 2,262 & 337 & 409 & 273.4 & \\ \hline

\multicolumn{6}{c}{Topic Classification} \\
\hdashline
TNEWS & 9,979 & 1,882 & 1,891 & 247.0 & \multirow{3}{*}{281.5} \\
Domain\_CLS & 9,941 & 1,271 & 1,288 & 248.2 & \\
CSL & 5,844 & 673 & 654 & 349.3 & \\ \hline

\multicolumn{6}{c}{Reading Comprehension with Common Sense} \\
\hdashline
C3 & 11,419 & 1,818 & 1,858 & 350.3 & \multirow{3}{*}{497.0} \\
LogiQA & 5,960 & 442 & 513 & 562.0 & \\
LogiQA2 & 10,320 & 1,132 & 1,225 & 578.6 & \\ \hline

\multicolumn{6}{c}{Common Sense} \\
\hdashline
E-KAR & 920 & 108 & 310 & 531.1 & \multirow{4}{*}{315.3} \\
Crisis & 2,239 & 202 & 279 & 294.2 & \\
CMOS & 9,994 & 3,221 & 3,215 & 203.0 & \\
COLD & 9,256 & 2,054 & 2,297 & 233.0 & \\ \hline

\multicolumn{6}{c}{Chinese Linguistic} \\
\hdashline
Metaphor & 3,443 & 381 & 367 & 300.0 & \multirow{3}{*}{311.8} \\
FCGEC & 9,264 & 282 & 485 & 312.4 & \\
NaCGEC & 5,095 & 637 & 637 & 322.9 & \\ \hline

\multicolumn{6}{c}{Named Entity Recognition} \\
\hdashline
CLUENER & 4,219 & 165 & 157 & 230.1 & \multirow{3}{*}{225.0} \\
MSRA\_NER & 9,788 & 143 & 2,643 & 233.9 & \\
Weibo\_NER & 678 & 117 & 118 & 211.0 & \\ \hline

\end{tabular}
\caption{Data Statistics for every used datasets. |Train|, |Dev|, and |Test| are the number of samples in different splits. Rat\_Len represents the average length of rationales in a dataset, while Rat\_len(task) is the average rationale length specific to each task.}
\label{app:dataset_combined}
\end{table*}

\clearpage
\clearpage

\section{Datasets and Tasks}\label{Appendix_datasets}
We collect 34 common Chinese NLU datasets across 11 tasks. Below is the introduction and source of each dataset in each task. The statistics of each dataset  after clustering and cleaning are in Table \ref{app:dataset_combined}. In addition, we provide the average length of generated rationales for each dataset and each task.

\subsection{Sentiment Analysis (Sentiment)}
Sentiment Analysis tasks  test the ability to assign appropriate sentiment labels for a given text. We include both sentence-level and aspect-level sentiment analysis tasks.  We use the following dataset:

\begin{itemize}
\item ChnSentiCorp: It stands for Chinese Sentiment Corpus. This sentence-level binary sentiment analysis task classifies the sentiment of a given text as either positive or negative. It's released in \cite{DBLP:journals/eswa/TanZ08}, and we obtain it from https://github.com/Tencent/TencentPretrain/wiki/Downstream-datasets. There is no mention of license or terms for use and/or distribution in this dataset.

\item SMP2020-EWECT: This sentence-level six-class sentiment analysis task classifies emotions in Weibo posts into one of six categories: positive, anger, sadness, fear, surprise, or no emotion. It's released as a competition dataset in https://smp2020ewect.github.io, and we obtain it from https://github.com/Tencent/TencentPretrain/wiki/Downstream-datasets. There is no mention of license or terms for use and/or distribution in this dataset.

\item ASAP\_ASPECT: This aspect-level ternary sentiment analysis task is based on restaurant reviews. Given a review and a specific aspect of the restaurant, the task aims to classify the sentiment as neutral, positive, or negative. It's released in \cite{DBLP:conf/naacl/BuRZYWZW21} and make publicly available in https://github.com/Meituan-Dianping/asap. We obtain the dataset from
https://aistudio.baidu.com/competition/detail/50/0/task-definition. This dataset is licensed under Apache-2.0.



\item BDCI\_Car: This aspect-level ternary sentiment analysis task involves car reviews from automotive forums. The objective is to determine whether the sentiment towards a given aspect of the car is positive, negative, or neutral. It's released as a competition dataset and are publicly available in https://www.datafountain.cn/competitions/310/datasets, where we obtain the dataset. There is no mention of license or terms for use and/or distribution in this dataset.

\end{itemize}

\subsection{Stance Detection (Stance)}
Stance Detection tasks  test the ability to identify the stance tendencies embedded in the text, typically one of support, opposition, and neutrality. We use the following dataset:
\begin{itemize}
\item NLPCC-stance: Given a Weibo post and the target it comments on, the task aims to classify the stance of the post as supportive, opposing, or neutral towards the target. It's released in \cite{DBLP:conf/nlpcc/XuZWGDX16} and make publicly available in https://huggingface.co/datasets/strombergnlp/nlpcc-stance, where we obtain the dataset. This dataset is licensed under Creative Commons Attribution 4.0.

\item DuReader\_yesno: This task involves polarity judgment in a reading comprehension context. Given a question and an answer, the task is to determine whether the answer expresses a positive or negative stance. It's released in \cite{DBLP:conf/acl/HeLLLZXLWWSLWW18} and made publicly available in https://github.com/baidu/DuReader. We obtain the dataset from https://www.luge.ai/luge/dataDetail?id=2. This dataset is licensed under Apache License, Version 2.0.

\item ReCO: It stand for Reading Comprehension Dataset on Opinion. Given a background, a question, and several options representing different viewpoints, the task is to select the correct viewpoint from the options in response to the question. The dataset is released in \cite{DBLP:conf/aaai/WangYZXW20} and made publicly available in https://github.com/benywon/ReCO, where we obtain the dataset. There is no mention of license or terms for use and/or distribution in this dataset.

\end{itemize}

\subsection{Natural Language Inference (NLI)}
Natural Language Inference tasks involve two input texts: a premise and a hypothesis. It tests the ability to to determine the logical relationship between the two texts, specifically whether the hypothesis is entailed by the premise, contradicts the premise, or is neutral to the premise.  We use the following dataset:
\begin{itemize}

\item OCNLI: It stands for Original Chinese Natural Language Inference. The Chinese texts are original and not translated. It's released in \cite{DBLP:conf/emnlp/HuRXLKM20} and make publicly available in https://github.com/cluebenchmark/OCNLI, where we obtain the dataset. This dataset is licensed under Attribution-NonCommercial 2.0 Generic (CC BY-NC 2.0).

\item CMNLI: This dataset translated from two English NLI datasets: XNLI \cite{DBLP:conf/emnlp/ConneauRLWBSS18} and MNLI \cite{DBLP:conf/naacl/WilliamsNB18}. It's publicly available in https://github.com/CLUEbenchmark/CLUE?tab=readme-ov-file, where we obtain the dataset.

\end{itemize}

\subsection{Paraphrase Detection (Paraphrase)}
The Paraphrase Detection task test the ability to determine the semantic similarity between pairs of texts.  We use the following dataset:

\begin{itemize}

\item LCQMC: It stands for Large-scale Chinese Question Matching Corpus. It is a paraphrase detection task, which is sourced  from the Baidu Zhidao domain. The input consists of two questions (Question 1 and Question 2), and the task is to judge whether the questions are semantically matching or not. It's released in \cite{DBLP:conf/coling/LiuCDZCLT18} and make publicly available in http://icrc.hitsz.edu.cn/Article/show/171.html. We obtain it from https://huggingface.co/datasets/C-MTEB/LCQMC. This dataset is licensed under a Creative Commons Attribution 4.0 International License

\item AFQMC: It stands for Ant Financial Question Matching Corpus. It is a question paraphrase detection task, which is sourced from  financial customer service. The input consists of two question texts (Text 1 and Text 2), and the task is to judge whether the questions are semantically matching or not. It's released in https://tianchi.aliyun.com/dataset/106411, where we obtain the datset. This dataset is licensed under a CC BY-NC 4.0.

\item BUSTM: It stands for XiaoBu Dialogue Short Text Matching. It is a question paraphrase detection task, which is sourced from  real conversations between users and intelligent assistants. The task is characterized by short, highly colloquial texts with challenging cases of high textual similarity but different semantics. The input consists of two questions (Question 1 and Question 2), and the task is to judge whether the questions are semantically matching or not. It's released in \cite{BUSTM} and make publicly available in https://github.com/CLUEbenchmark/FewCLUE/tree/main/datasets/bustm. We obtain the dataset from https://huggingface.co/datasets/suolyer/bustm. There is no mention of license or terms for use and/or distribution in this dataset.

\item PAWS: It stands for Paraphrase Adversaries from Word Scrambling. It's a text pair paraphrase detection task. This task is characterized by high lexical overlap. Given two texts (Text 1 and Text 2), the task aims to determine if they have the same meaning. It's released in \cite{DBLP:conf/emnlp/YangZTB19} and made publicly available in https://github.com/google-research-datasets/paws. We obtain the dataset from https://huggingface.co/datasets/google-research-datasets/paws-x and extract the Chinese part from this multilingual dataset. This dataset could be freely used for any purpose.

\item QBQTC: It stands for QQ Browser Query Title Corpus. It's a query paraphrase detection task, which is sourced from real user queries in QQ search engines.  The input consists of two queries (Query 1 and Query 2), and the task is to determine the degree of relevance, outputting one of the following: low relevance, somewhat relevant, or highly relevant. It's released and made publicly available in https://github.com/CLUEbenchmark/QBQTC. We obtain the dataset from  https://huggingface.co/datasets/C-MTEB/QBQTC. There is no mention of license or terms for use and/or distribution in this dataset.
\end{itemize}

\subsection{Coreference Resolution (Coreference)}
Coreference Resolution tasks  test the ability to identify expressions of the same entity in some given text. We use the following dataset:

\begin{itemize}
\item CLUEWSC2020: Given a sentence containing a pronoun and a noun (or noun phrase) in the sentence, the task is to determine whether or not the noun (or noun phrase) is the object to which the pronoun refers. It's released and made publicly available in https://github.com/CLUEbenchmark/CLUEWSC2020. We obtain the dataset from  https://huggingface.co/datasets/wyp/clue-wsc. There is no mention of license or terms for use and/or distribution in this dataset.
\end{itemize}

\subsection{Reading Comprehension (RC)}
Reading Comprehension tasks test the ability to extract answer to a question given a passage that contains the answer. We use the following datasets:

\begin{itemize}
\item CMRC2018: This span-based reading comprehension task is to extract a continuous sequence from the given passage that best answers the given question. It's released in \cite{DBLP:conf/emnlp/CuiLCXCMWH19} and made publicly available in https://github.com/ymcui/cmrc2018. We obtain the dataset from https://huggingface.co/datasets/hfl/cmrc2018. It's licensed under Creative Commons Attribution Share Alike 4.0.

\item DuReader\_Robust: This span-based reading comprehension task is to extract a continuous sequence from the given passage that best answers the given question. It is characterized by passage that contains close but wrong answers. It's released in \cite{DBLP:conf/acl/TangL0H0020} and made publicly available in https://github.com/baidu/DuReader. We obtain the dataset from https://www.luge.ai/luge/dataDetail?id=1. It's licensed under the Apache License, Version 2.0.

\item DuReader\_checklist: In this span-based reading comprehension task, given a question and a passage, you first need to determine whether the passage contains an answer to the given question. If it does, you need to extract a continuous sequence from the passage that best answers the question; otherwise, you need to output "no answer.". It's released in \cite{DBLP:conf/acl/HeLLLZXLWWSLWW18} and made publicly available in https://github.com/baidu/DuReader. We obtain the dataset from https://www.luge.ai/luge/dataDetail?id=3. It's licensed under the Apache License, Version 2.0.

\end{itemize}

\subsection{Topic Classification (Topic)}
Topic Classification tasks test the ability to classify texts into predefined categories based on their content. We use the following datasets:
\begin{itemize}
\item TNEWS: It stands for TouTiao Text Classification for News Titles. The predefined categories include: story, culture, entertainment, sports, finance, home, automobile, education, technology, military, travel, world, stock, agriculture, and games. Each news item belongs to one and only one category. It's released and made publicly available in https://github.com/aceimnorstuvwxz/toutiao-text-classfication-dataset. We obtain the dataset from https://huggingface.co/datasets/C-MTEB/TNews-classification. There is no mention of license or terms for use and/or distribution in this dataset.

\item Domain\_CLS: It's a text domain classification task. The categories include: transportation and warehousing, accommodation and catering, information software, agriculture, manufacturing, healthcare, international organizations, construction, real estate, government organizations, education, cultural and sports entertainment, water conservancy and environment, electricity and gas production, science and technology, leasing and legal services, mining, and finance. Each text belongs to one and only one domain. It's released and made publicly available in https://modelscope.cn/datasets/iic/nlp\_domain\_classification\_chinese\_testset/files, where we obtain the dataset. It's licensed under the Apache License, Version 2.0.

\item CSL: It stands for Chiense Scientific Literature. It is a academic paper title classification task. The categories include: law, engineering, economics, science, agriculture, management, medicine, history, arts, philosophy, education, military science, and literature. Each title belongs to one and only one discipline. It's released 
\cite{DBLP:conf/coling/LiZ0S0MZ22} and made publicly available in https://github.com/ydli-ai/CSL, where we obtain the dataset. It's licensed under the Apache License, Version 2.0.

\end{itemize}

\subsection{Reading Comprehension with Common Sense (RC. w/ CS)}
Reading Comprehension with Common Sense tasks test the ability to comprehend the background information and use common sense reasoning to select the most appropriate answer from the provided options. We use the following datasets:

\begin{itemize}
\item $C^3$: It is a multiple-choice reading comprehension task. Given a background (which could be a long text or a dialogue), a question, and some options (2-4), you need to understand the background and choose the correct answer from the four options based on the question. It's released in \cite{DBLP:journals/tacl/SunYYC20} and made publicly available in https://dataset.org/c3. We obtain it from https://huggingface.co/datasets/dataset-org/c3. There is no mention of license or terms for use and/or distribution in this dataset.

\item LogiQA: It is a multiple-choice reading comprehension task with logical reasoning. Given a background text, a question, and four options, you need to perform logical reasoning to choose the most appropriate answer. It's released in \cite{DBLP:conf/ijcai/LiuCLHWZ20} and made publicly available in https://github.com/lgw863/LogiQA-dataset, where we obtain the dataset. There is no mention of license or terms for use and/or distribution in this dataset.

\item LogiQA2: It is an extended version of LogiQA. Given a background text, a question, and four options, you need to perform logical reasoning to choose the most appropriate answer. It's released in \cite{DBLP:journals/taslp/LiuLCTDZZ23} and made publicly available in https://github.com/csitfun/LogiQA2.0, where we obtain the dataset. This dataset is licensed under a Creative Commons Attribution-NonCommercial-ShareAlike 4.0 International License.

\end{itemize}

\subsection{Common Sense}
Common Sense tasks test the ability to understand and reason based on general knowledge and societal norms to analyze the given text. Our datasets include determining the usefulness of information for emergency response, identifying moral content, detecting discriminatory language, and recognizing word analogies. We use the following datasets:

\begin{itemize}
\item E-KAR: It stands for Explainable Knowledge-intensive Analogical Reasoning. It's a multiple-choice word analogy recognition task. Given a problem (a set of words) and several options (each option is also a set of words), you need to analyze the relationship implied between the words in the problem and each option, then choose the answer from the options that most closely matches the relationship in the problem. It's released in \cite{DBLP:conf/acl/ChenXFSLZSLXZ22} and made publicly available in https://ekar-leaderboard.github.io. We obtain it from https://huggingface.co/datasets/jiangjiechen/ekar\_chinese. This dataset is licensed under CC BY NC SA 4.0.

\item Crisis: This ternary classification task is to determine the informational value in the blog post for emergency response or rescue efforts. Given a blog post, you need to classify it as useful, not useful, or indeterminate. It's released and made publicly available in https://github.com/lzquancumtb/CrisisNLP-C, where we obtain the dataset. There is no mention of license or terms for use and/or distribution in this dataset.

\item CMOS: It stands for Chinese MOral Sentence. This binary classification task to classify whether the morality implied in the text is positive or negative.  It's released in and made publicly available in https://github.com/blcunlp/Chinese-MOral-Sentence-Dataset, where we obtain the dataset. This dataset is licensed under Creative Commons Attribution 4.0 International License. 

\item COLD: It stands for Chinese Offensive Language Detection. This binary classification task is to determine whether the given text contains discriminatory or offensive meaning towards the given theme. If the text exhibits discrimination on the specific theme, the output is "yes"; otherwise, the output is "no." \textcolor{red}{It contains content that may be profane, vulgar, or offensive.} It's released in \cite{DBLP:conf/emnlp/DengZ0ZMMH22} and made publicly available in https://github.com/thu-coai/COLDataset. We obtain the dataset from https://huggingface.co/datasets/thu-coai/cold. This dataset is licensed under Apache License, Version 2.0.

\end{itemize}

\subsection{Chinese Linguistic (Linguistic)}
Chinese Linguistic tasks test the ability specifically related to Chinese linguistic, such as recognizing metaphors and grammatical errors. We use the following datasets: 

\begin{itemize}
\item Metaphor: This ternary classification task is to identify metaphors in Chinese involving verbs and nouns aims to recognize whether a text uses metaphors with verbs or nouns. If no metaphor is used, the output should be "no metaphor." If a metaphor is present, you need to further identify whether the text uses a verb metaphor or a noun metaphor. It's released and made publicly available in https://github.com/DUTIR-Emotion-Group/CCL2018-Chinese-Metaphor-Analysis, where we obtain the dataset. There is no mention of license or terms for use and/or distribution in this dataset.

\item FCGEC: It stands for Fine-Grained Corpus for Chinese Grammatical Error Correction. Given a text, you need to check whether it contains any grammatical errors. The types of grammatical errors in this task include: improper word order, improper collocation, missing components, redundant components, structural confusion, illogicality, and unclear semantics. If there are no grammatical errors, the output should be "no error.". Note that there is at most one type of grammatical error in a text. It's released in \cite{xu2022fcgec} and made publicly available in https://github.com/xlxwalex/FCGEC, where we obtain the dataset. This dataset is licensed under Apache License, Version 2.0.

\item NaCGEC: It stands for Native Chinese Grammatical Error Correction. Given a text, you need to check whether it contains any grammatical errors. The types of grammatical errors in this task include: improper collocation, illogicality, improper word order, missing components, redundant components, and mixed sentence patterns. If there are no grammatical errors, the output should be "no error." Note that there is at most one type of grammatical error in a text. It's released in \cite{ma-etal-2022-linguistic} and made publicly available in https://github.com/masr2000/NaCGEC, where we obtain the dataset. This dataset is licensed under Apache License, Version 2.0.

\end{itemize}

\subsection{Named Entity Recognition (NER)}
Named Entity Recognition (NER) tasks test the ability to identify and classify entities within a text into predefined categories such as names of people, organizations, locations, dates, and other proper nouns. We use the following datasets:

\begin{itemize}
\item CLUENER: This task aims to identify entities in the given text and classify them into: address, book title, company, game, government, movie, name, organization, scenic spot, and position. It's released in \cite{DBLP:journals/corr/abs-2001-04351} and made publicly available in https://github.com/CLUEbenchmark/CLUENER2020. We obtain the dataset from https://huggingface.co/datasets/xusenlin/clue-ner. This dataset is licensed under Apache License, Version 2.0.

\item MSRA\_NER: This task aims to  identify entities in the given text and classify them into: location, person, and organization. It's released in \cite{DBLP:conf/acl-sighan/Levow06}. We obtain the dataset from https://github.com/Tencent/TencentPretrain/wiki/Downstream-datasets. There is no mention of license or terms for use and/or distribution in this dataset.

\item Weibo\_NER: This task aims to  identify entities in the given text from Chinese social media (Weibo) and classify them into: location, person, organization, and geopolitical. It's released in \cite{DBLP:conf/emnlp/PengD15} and made publicly available in https://github.com/hltcoe/golden-horse, where we obtain the dataset. This dataset is licensed under Creative Commons Attribution-ShareAlike 3.0 Unported License (CC BY-SA 3.0).

\end{itemize}

\section{Chinese Cases}
We provide the original Chinese versions of all previous case studies, along with the Chinese scoring criteria and human evaluation interface.

\begin{CJK}{UTF8}{gbsn}

\begin{table*}[t]
\centering
\small
\begin{tabular}{p{15cm}}  
\hline
(Original Prompt)\newline
\textbf{Instruction}: 你是一个完成问题对匹配任务的专家。该问题对匹配任务来源于百度知道领域，旨在理解并比较两个中文问题的语义内容是否一致。输入包含两个问题（问题1和问题2），你需要判断这两个问题是否语义相似或表达相同的查询意图。如果两个问题在本质上询问的是相同的信息，那么他们的语义关系为匹配；否则，语义关系为不匹配。 \newline \newline
现在，阅读问题1和问题2 \newline 
\textbf{Input}: \newline 问题1: 百合是什么意思？ \newline 问题2: 百合什么意思?  \newline \newline
然后，一步一步思考，先给出推理过程，最后给出它们之间的关系作为答案。\newline \newline
\textbf{Rationale}: \footnotesize  \textcolor{red}{1. 首先,} 观察问题1和问题2的文字表述，可以看到两者的主要区别在于“是”和“什么”的位置。  从语义上看，“百合是什么意思”和“百合什么意思”都在询问“百合”这个词的含义。尽管两者的词序略有不同，但它们的核心意图是一致的，都是在寻求对“百合”这个词的解释。 \newline  因此，得出答案：匹配\\ \hline

\end{tabular}
\caption{Original Chinese case in Table \ref{prompt_rat_rationale_original}.}
\label{prompt_rat_rationale_original_zh}
\end{table*}

\begin{table*}[t]
\centering
\small
\begin{tabular}{p{15cm}}  
\hline
(Prompt with \textbf{Label})  \newline
 \textbf{Instruction} \newline 
现在，阅读问题1、问题2和它们之间的关系 \newline 
\textbf{Input} \newline
\textbf{关系}: 匹配 \newline 
然后，一步一步思考，给出得到该关系的推理过程。\newline\newline
\textbf{Rationale}: \footnotesize 1. 首先，观察问题1“百合是什么意思”和问题2“百合什么意思”，可以看出这两个问题的文字内容几乎完全一致，只是“是”和“什么”的位置有所调换。这种微小的差异并不影响两个问题的核心语义。 \newline 其次，两个问题都在询问“百合”的定义或解释，意图是相同的，即想要了解“百合”这个词的含义。 \newline 3. 综合以上两点，可以判断问题1和问题2在语义上是相似的，表达的是相同的查询意图。 \newline  因此，得出答案：匹配。 \\ \hline 
\end{tabular}
\caption{Original Chinese case in Table \ref{prompt_rat_rationale_label}. }
\label{prompt_rat_rationale_label_zh}
\end{table*}

\begin{table*}[t]
\centering
\small
\begin{tabular}{p{15cm}} 
\hline
(Prompt with \textcolor{red}{Wrong} \textbf{Label})  \newline
 \textbf{Instruction} \newline 
现在，阅读问题1、问题2和它们之间的关系 \newline 
\textbf{Input} \newline
\textbf{关系}: \textcolor{red}{不匹配} \newline 
然后，一步一步思考，给出得到该关系的推理过程。\newline \newline 
\textbf{Rationale}: \footnotesize 1. 首先，我们分析问题1“百合是什么意思”，这是在询问“百合”一词的含义或定义。接下来，我们分析问题2 “百合什么意思”，它同样是在询问 “百合”一词的含义或定义。\textcolor{red}{乍一看，这两个问题似乎问的是相同的内容。但是，}为了确保语义的准确性，我们需要考虑两个方面：一个是问题的表达方式，另一个是问题背后的意图。\newline
2. 在表达方面，问题1使用了“是什么意思”，而问题2则简化为“什么意思”。虽然表达方式略有不同，但它们的意图是相同的，即询问一个词的含义。然而，语义匹配不仅取决于问题的表面表达，还取决于提出问题的具体语境和环境。\textcolor{red}{即使问题表面上看起来询问的是相同的意思，但在某些情况下，比如在更广泛的语境中，可能需要更精确的表达来区分不同的查询意图}。 \newline
3. 因此，在这种情况下，如果我们考虑到这两个问题在表达方式上的细微差别，以及在某些特定语境中可能需要更明确的表达方式，我们就可以得出结论：虽然这两个问题询问的是相同的信息，但它们在表达方式上的差异导致它们在语义上并不完全匹配。
\newline
 因此，得出答案：不匹配。

\\ \hline 
\end{tabular}
\caption{Original Chinese case in Table \ref{prompt_rat_rationale_labelwrong}. }
\label{prompt_rat_rationale_labelwrong_zh}
\end{table*}

\begin{table*}[t]
\centering
\small
\begin{tabular}{p{15cm}}  
\hline
(Prompt with \textbf{Label} \& \textbf{8 Exemplars}) \newline
\textbf{Instruction} \newline \newline 
\textbf{[示例1]} \newline \footnotesize
问题1: 求小说电子书 问题2: 求《甄嬛》小说电子书！  关系: 不匹配
\newline Rationale: 1. 分析问题1的关键信息：问题1中的关键信息是“小说电子书”，这是一个非常宽泛的请求，没有指定具体的小说名称或类型，因此可以理解为求任何一本小说的电子书版本。2. 分析问题2的关键信息：问题2中的关键信息是“《甄嬛》小说电子书”，这是一个具体的请求，指定了小说的名称为《甄嬛》，因此求的是特定小说《甄嬛》的电子书版本。3. 比较两个问题的关键信息：问题1和问题2的请求对象不同。问题1请求的是任意小说的电子书，而问题2请求的是特定小说《甄嬛》的电子书。4. 判断语义内容是否一致：虽然两个问题都在求小说的电子书版本，但问题1没有指定具体的小说，而问题2指定了《甄嬛》这本小说。因此，两个问题在语义内容上不一致。5. 综合分析得出结论：根据以上分析，问题1和问题2在询问的信息上有本质的不同，问题1是一个泛泛的请求，而问题2是一个具体的请求，两个问题的查询意图不相同。因此，得出答案: 不匹配。\newline
\textbf{[示例2]}... \newline 
\newline
现在，阅读问题1、问题2和它们之间的关系 \newline
\textbf{Input} \newline
\textbf{关系}: 匹配 \newline 
然后，参照上述示例，给出得到该关系的推理过程。\newline\newline
\textbf{Rationale}: \footnotesize \textcolor{red}{1. }分析问题1的关键词和意图：问题1中的关键词是“百合”和“是什么意思”，意图是询问“百合”这个词的含义或者解释。\newline 
\textcolor{red}{2. } 分析问题2的关键词和意图：问题2中的关键词也是“百合”和“什么意思”，意图同样是询问“百合”这个词的含义或者解释。\newline 
\textcolor{red}{3. } 对比两个问题的语义内容：尽管问题1和问题2使用了不同的词汇， \textcolor{red}{（“是什么意思”和“什么意思”）}, 但它们都集中在询问“百合”这个词的含义。\newline 
\textcolor{red}{4. } 判断两个问题是否表达相同的查询意图：由于两个问题都在寻求对“百合”这个词的解释或含义，\textcolor{red}{即使是通过略微不同的表述方式，} 它们的查询意图是一致的。\newline 
\textcolor{red}{5. } 综合判断两个问题的匹配关系：结合上述所有信息，问题1和问题2在本质上询问的是相同的信息，即“百合”这个词的含义。 \newline
因此，得出答案：匹配。
\\ \hline
\end{tabular}
\caption{Original Chinese case in Table \ref{prompt_rat_rationale_labelexem}.}
\label{prompt_rat_rationale_labelexem_zh}
\end{table*}

\begin{table*}[t]
\centering
\small
\begin{tabular}{p{15cm}}  
\hline
(Prompt with \textbf{Label} \& \textbf{Label Criteria})  \newline
\textbf{Instruction} \newline \newline
\textbf{Label Criteria:} \newline \footnotesize
匹配: 两个问题的匹配关系为是，如果满足以下任一条件：
1. 相同意图：两个问题明确地表达了相同的查询意图，即使它们的字面表达或细节描述有所不同。
2. 核心查询相同：尽管可能包含一些边缘信息的差异，但两个问题的核心查询对象或目的相同。
3. 可互换性：一个问题的答案能够合理地作为另一个问题的答案，即这两个问题在实际应用场景中可以互换。
\newline
不匹配：两个问题匹配关系为否，如果满足以下任一条件：
1. 不同意图：两个问题表达了不同的查询意图，它们询问的是根本不同的信息或事物。
2. 核心查询不同：即使两个问题在某些方面有相似性，它们的核心关注点或主要查询目标有所差异。
3. 答案不可互换：一个问题的答案不能合理地作为另一个问题的答案，表明这两个问题在意图或需求上有本质的不同。 \newline 
\textbf{Input} \newline
\textbf{关系}: 匹配 \newline 
然后，一步一步思考，给出得到该关系的推理过程。\newline\newline
\textbf{Rationale}: \footnotesize \textcolor{red}{1. }首先，我们来分析问题1：“百合是什么意思”和问题2：“百合什么意思”。从字面上看，这两个问题的表述非常相似，只是“是”和“什么”的位置互换了。这种差异并不影响问题的核心查询意图。\newline
\textcolor{red}{2. }其次，从查询意图来看，两个问题都在询问“百合”这个词的含义。无论是，\textcolor{red}{“百合是什么意思”还是“百合什么意思”}, 都是在寻求对“百合”这个词的解释或定义。
\newline
\textcolor{red}{3. }再者，考虑到答案的可互换性，如果一个问题得到了解答，那么另一个问题的答案也是相同的。例如，如果“百合是什么意思”的答案是“百合是一种植物”，那么“百合什么意思”的答案也是“百合是一种植物”。\newline
因此，综合以上分析，这两个问题的语义关系是匹配的。因此，得出答案：匹配。
\\ \hline
\end{tabular}
\caption{Original Chinese case in Table \ref{prompt_rat_rationale_labelcriteria}.}
\label{prompt_rat_rationale_labelcriteria_zh}
\end{table*}

\begin{table*}[t]
\centering
\small
\begin{tabular}{p{2cm} p{12.5cm}}
\hline
\textbf{Metrics}           & \textbf{Scoring Criteria}         \\ \hline
简洁性       & \begin{tabular}[c]{@{}p{12.5cm}@{}} 
\textbf{5分}: 解释非常简洁，没有不必要的细节或重复；每个词都增加了价值。\\ 
\textbf{4分}: 解释大致简洁，但有些许冗余或细节略显过多。\\ 
\textbf{3分}: 解释中等简洁；包含一些不必要的句子或重复，应该去除。\\ 
\textbf{2分}: 解释冗长，有显著的冗余或无关的细节，影响清晰度。\\ 
\textbf{1分}: 解释过于冗长或重复，完全掩盖了要点。
\end{tabular} \\ \hline
全面性 & \begin{tabular}[c]{@{}p{12.5cm}@{}} 
\textbf{5分}: 解释全面覆盖所有重要方面和相关背景信息；没有遗漏关键内容。\\ 
\textbf{4分}: 解释涵盖了大部分关键方面，但有些细节略显不足或有小幅遗漏。\\ 
\textbf{3分}: 解释涵盖了一些关键方面，但缺少重要元素或关键部分缺乏背景信息。\\ 
\textbf{2分}: 解释不完整，遗漏了多个关键点或必要的背景信息，影响清晰度。\\ 
\textbf{1分}: 解释未能回答核心问题；缺失的关键内容使解释无效。
\end{tabular} \\ \hline
连贯性         & \begin{tabular}[c]{@{}p{12.5cm}@{}} 
\textbf{5分}: 解释遵循一个完全连贯、一步步展开的逻辑结构，没有矛盾或推理跳跃。\\ 
\textbf{4分}: 解释大致逻辑清晰，但偶尔有小的连贯性问题或步骤感觉突然或脱节。\\ 
\textbf{3分}: 解释在推理过程中有明显的间隙或不一致，但整体方向仍然明确。\\ 
\textbf{2分}: 解释结构较差，频繁出现逻辑间隙、矛盾或进展不清晰。\\ 
\textbf{1分}: 解释完全缺乏逻辑连贯性，推理杂乱无章或脱节。
\end{tabular} \\ \hline
忠实性      & \begin{tabular}[c]{@{}p{12.5cm}@{}} 
\textbf{5分}: 解释完全忠实于输入和标签，完全基于证据，没有偏离。\\ 
\textbf{4分}: 解释大致忠实，但包括一些轻微的不准确或无关点，且没有显著影响。\\ 
\textbf{3分}: 解释有中等程度的忠实性；与输入和标签部分一致，但有明显的不准确之处。\\ 
\textbf{2分}: 解释与输入或标签偏离较大，存在频繁的不准确或无关的推理。\\ 
\textbf{1分}: 解释与输入或标签不符，理由不正确或编造。
\end{tabular} \\ \hline
多样性         & \begin{tabular}[c]{@{}p{12.5cm}@{}} 
\textbf{5分}: 解释非常多样，展现了不同的推理风格、措辞或视角。\\ 
\textbf{4分}: 解释大多样，但在不同示例中有一些重复模式或相似措辞。\\ 
\textbf{3分}: 解释表现出中等多样性；存在一定的变化，但显著的重复性也比较明显。\\ 
\textbf{2分}: 解释缺乏多样性，推理风格或措辞重复，主导了回答。\\ 
\textbf{1分}: 解释完全重复，推理、风格或措辞没有任何变化。
\end{tabular}   \\ \hline                                        
\end{tabular}
\caption{Original Chinese scoring criteria in Table \ref{app:scoring_criteria}.}
\label{app:scoring_criteria_zh}
\end{table*}

\clearpage
\clearpage

\begin{figure*}[t]
\centering
\includegraphics[scale = 0.3]{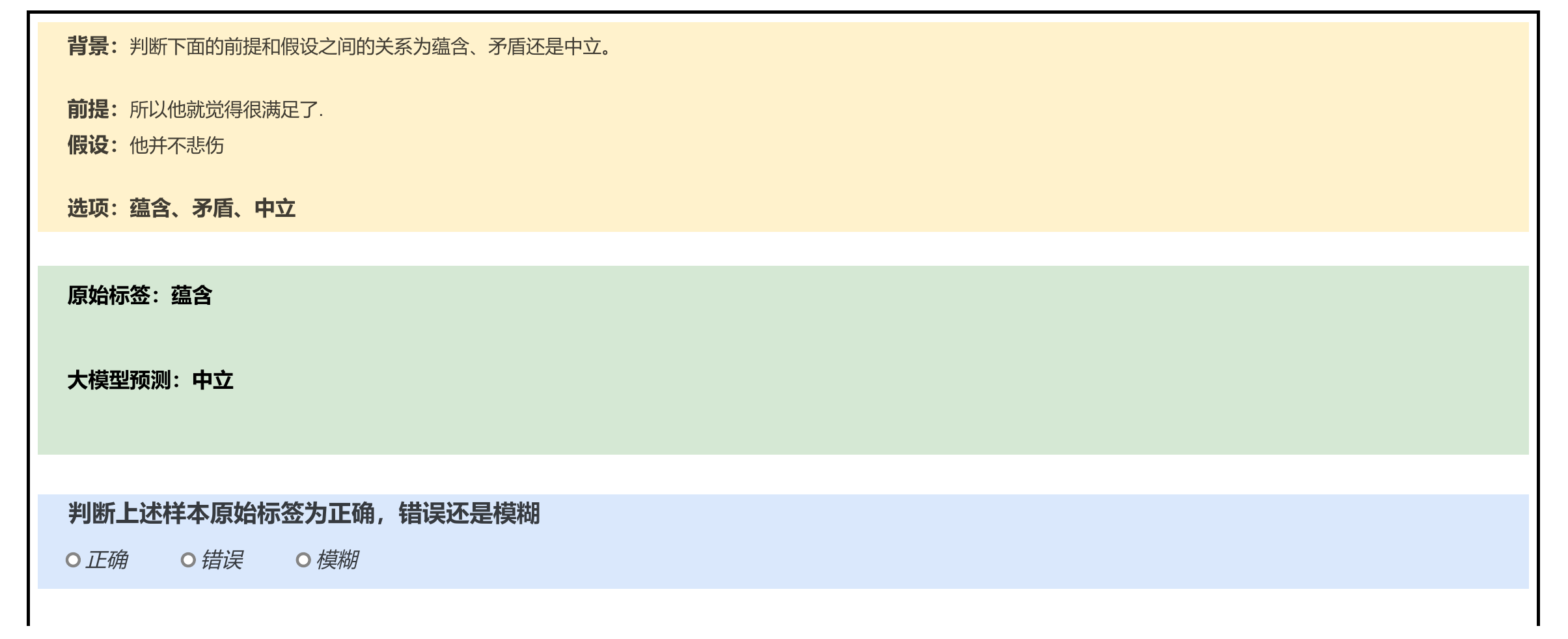}
\caption{Screenshot of original user interface for \textbf{judging label accuracy} in Figure \ref{screenshot_cor}.} 
\label{screenshot_cor_zh}
\end{figure*}

\begin{figure*}[t]
\centering
\includegraphics[scale = 0.3]{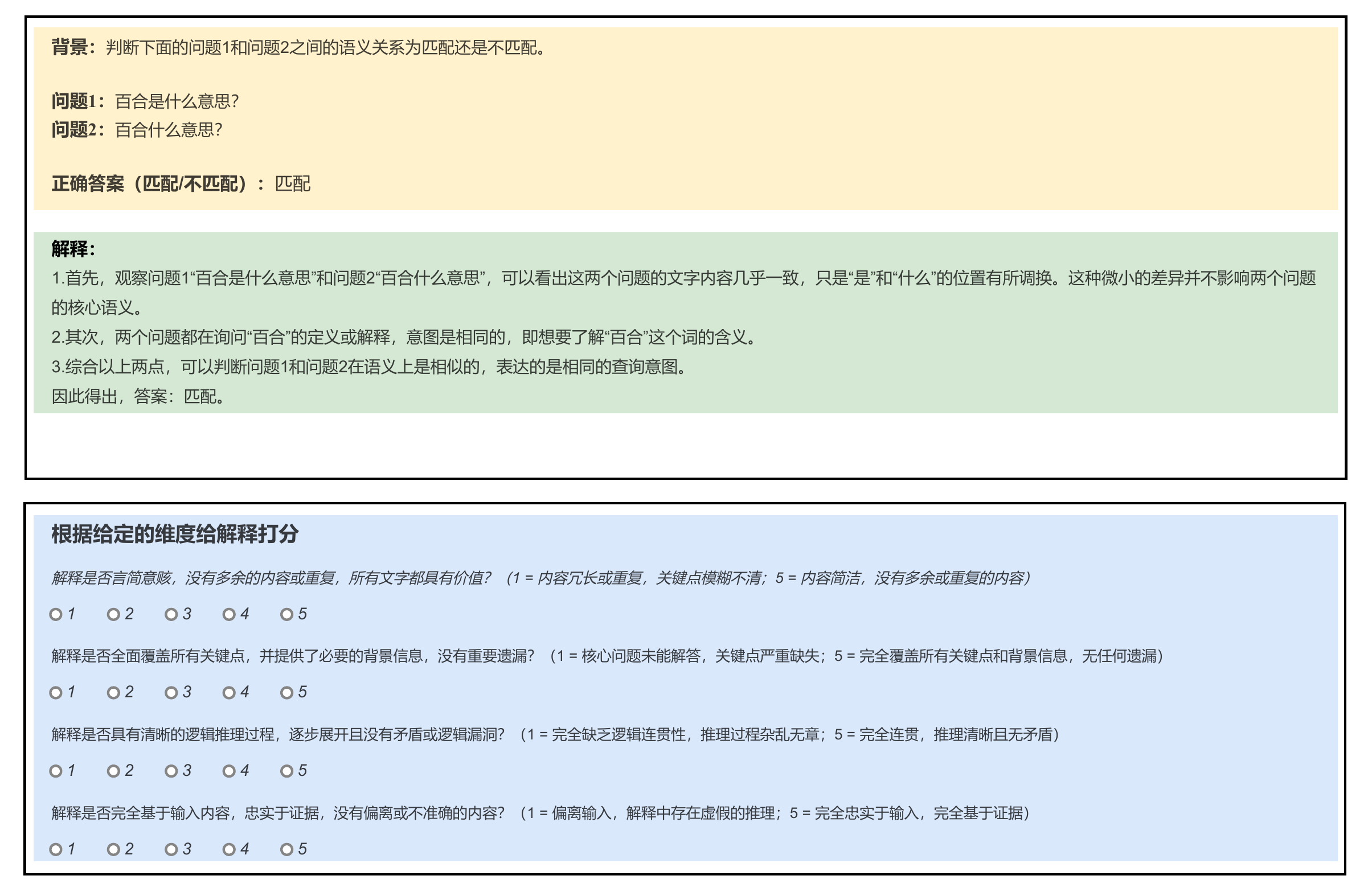}
\caption{Screenshot of original user interface for  \textbf{scoring rationale quality} in Figure \ref{screenshot_dimension}.} 
\label{screenshot_dimension_zh}
\end{figure*}

\begin{figure*}[t]
\centering
\includegraphics[scale = 0.3]{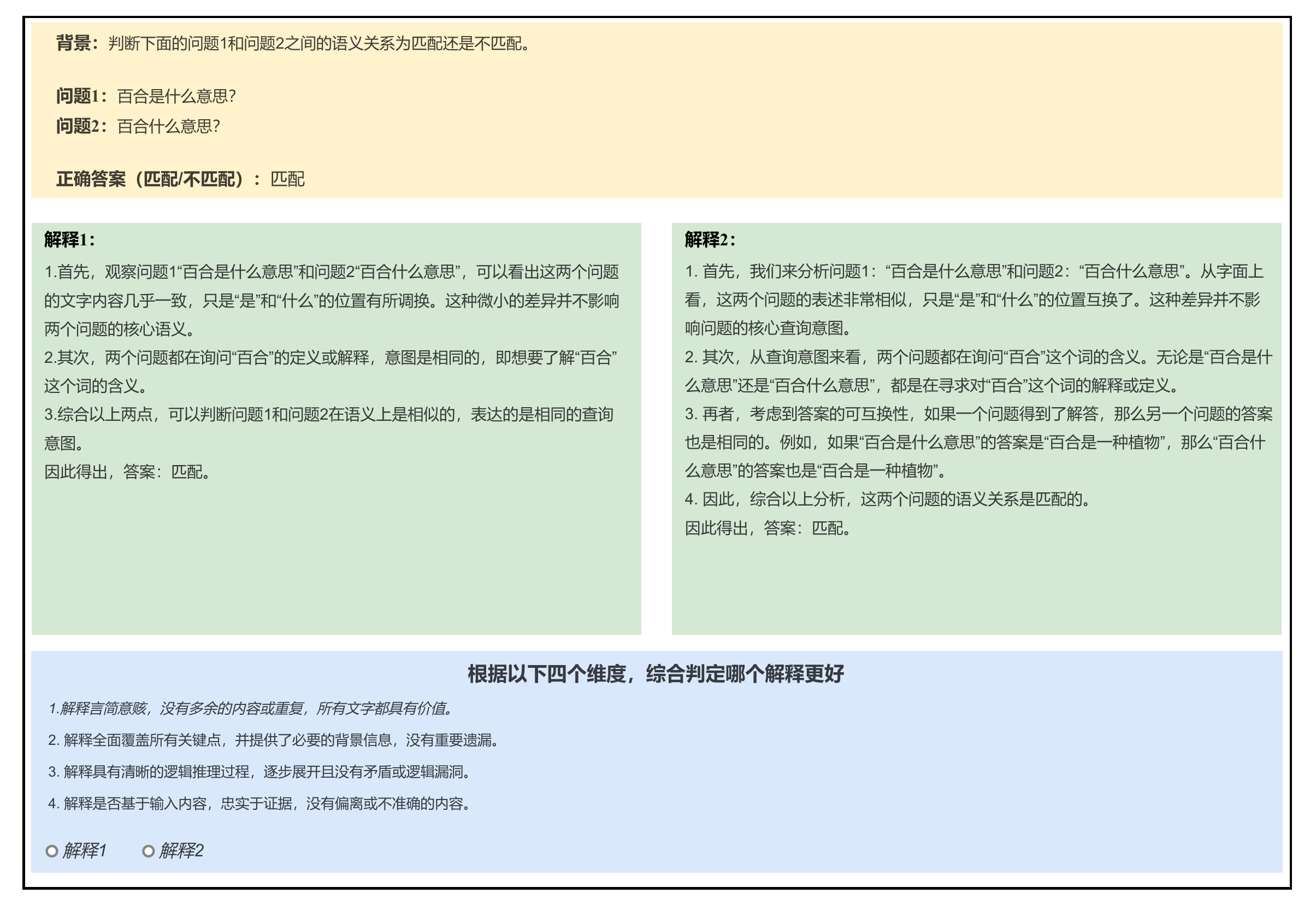}
\caption{Screenshot of original user interface for \textbf{judging  rationale comparison} in Figure \ref{screenshot_judge_qua}.} 
\label{screenshot_judge_qua_zh}
\end{figure*}

\begin{table*}[t]
\centering
\small
\begin{tabular}{p{1cm}p{13.5cm}}  
\hline
Method & \textbf{Instruction Prompt}: \newline 输入: 问题1: ; 问题2: ;\newline 判断下面的问题1和问题2之间的语义关系。 \\ \hline

Label-Only & (输入: ...; 判断 ...;) 直接输出匹配或者不匹配作为答案。 \\ \hline
\textbf{Reason}& (输入: ...; 判断 ...;) 先给出推理过程，结尾以“因此得出，答案：”给出匹配或者不匹配作为答案。\\ \hline
\textbf{Explain} & (输入: ...; 判断 ...;) 先直接输出匹配或者不匹配作为答案，然后给出得到该答案的推理过程。 \\
\hline
\end{tabular}
\caption{Original Chinese Instruction Prompt in Table \ref{Prompt_methods} }
\label{Prompt_methods_zh}
\end{table*}

\begin{table*}[t]
\centering
\begin{tabular}{p{2.5cm}p{12.5cm}}
\hline
\textbf{Error Type}         & \textbf{Definition}                                                                                                                                               \\ \hline
\textbf{理解错误}                & 当模型在解释过程中未能应用普遍的世界知识、日常推理或公认事实时，就会发生理解错误。这包括误解隐含含义、讽刺或文化引用，导致与基本现实原则相矛盾的解释。这类错误通常源于对社会规范、物理现实或典型因果关系的误解，而这些是人类读者可以自然理解的。 
\\ \hline
\textbf{逻辑错误} & 当模型用于解释自然语言输入的推理过程或策略存在缺陷时，就会发生逻辑错误。这包括错误地应用逻辑原则，如推理错误、归纳不当，或未能正确理解因果关系。即使模型正确理解了语言，如果由于错误的假设、不当的泛化或推理链中的矛盾，推理过程仍可能偏离，导致逻辑错误。                                 \\ \hline
\textbf{上下文错误} & 当模型未能正确解释或应用文本或话语中的相关上下文信息时，就会发生上下文错误。这可能包括忽略前文句子中的关键信息、在对话中忽略对话历史，或未能整合跨句子的引用和隐含信息。                            \\ \hline
\textbf{语言错误}      & 语言错误涉及语法、词汇、句法或语义解释方面的错误。这类错误发生在模型误解句子结构或未能理解词语和短语的组合方式时。例如，在复杂或嵌套句中，模型可能难以正确识别主语、动词和修饰语之间的关系。这类错误通常源于解析问题或对语法规则的误解。                \\ \hline
\end{tabular}
\caption{Original Chinese Difenitions of Error Types}
\label{Error_type_def_zh}
\end{table*}

\begin{table*}[]
\centering
\small
\begin{tabular}{p{1.6cm}p{5.2cm}p{7.8cm}}  
\hline
Task & Instruction \& Text   & Reasoning Process \& Answer     \\ \hline

Stance \newline NLPCC  & \textbf{Instruction}: 判断下面的文本对于给定评论对象的立场 \newline \textbf{文本}: 支持，之前被电动车在人行道撞过。 \newline \textbf{评论对象}: 禁摩限电 \newline \textbf{答案}: 支持 & \footnotesize 1. 首先，我们分析一下文本内容。文中提到 “支持”，说明\textcolor{red}{作者支持电动自行车在人行道上行驶的行为}。2. 接下来，我们需要将这一立场与评论的主题联系起来。禁令的目的是限制电动自行车在城市中的通行，以保护行人和车辆的安全。文中提到的支持电动自行车在人行道上行驶与禁止电动自行车的目的是相悖的。3. 由于文中提到的支持立场与评论的目标相悖，我们可以推断作者对评论对象持反对立场。\newline 因此，得出答案： \textcolor{red}{反对}。
\\ \hline
RC. w/CS. \newline C3  & \textbf{Instruction}: 阅读下面的背景和问题，从选项中选择出最合适的答案。 \newline
\textbf{背景}：女人： 你们公司通常几点下班？男人: 星期四是4:30，其他时间是5:00。如果周末加班，我就不清楚了。\newline \textbf{问题}：男人星期六几点下班？ \newline \textbf{选项}: A. 5:00; B. 5:30; C. 4:30; D. 不确定 \newline \textbf{回答}: D. 不确定 & \footnotesize 1.首先，我们从对话中提取关键信息。该男子提到：“星期四是 4:30，其他日子是 5:00。”这表明，除了星期四，通常工作日都是 5:00 结束。然后他补充说：“如果周末有加班，我不确定。”这意味着周末加班时间不确定。2. 接下来，我们分析这个问题： “这位男士周六几点下班？” \textcolor{red}{因为他没有明确指出周六的工作时间，而只提到了加班情况下的不确定性，所以我们可以推断，如果没有加班，周六的工作时间是固定的}。鉴于其他工作日的下班时间都是 5:00，我们可以合理地推断出，如果不加班，周六的下班时间也是 5:00。\newline 因此，答案是 \textcolor{red}{A. 5:00}. 
\\ \hline
\end{tabular}
\caption{Original Chinese cases in Table \ref{Reason_Reason_Wrong_Case}.}
\label{Reason_Reason_Wrong_Case_zh}
\end{table*}

\begin{table*}[]
\centering
\small
\begin{tabular}{p{1cm}p{14.6cm}}  
\hline
Model & \textbf{背景：} 在上海闸北区逛一逛，到处都可以找到您心仪的茶叶。上海不产茶叶，却是中国最大的茶叶消费与流通城市。\newline \textbf{问题：} 请选出与试题内容一致的一项。 \newline \textbf{选项：} A. 上海的茶多销往国外；B. 上海闸北区种植茶叶；C. 上海的茶叶消费量很大；D. 上海是中国最大的茶叶产地 \textbf{答案：} . 上海的茶叶消费量很大 \\ \hline
Hunyuan & \footnotesize 1.首先，我们需要仔细阅读背景信息，理解其内容。背景提到，上海虽然不产茶叶，却是中国最大的茶叶消费与流通城市。这提供了两个关键信息：上海的茶叶消费量大，且茶叶流通活跃。\newline
2.接下来，我们逐个分析选项：
A. 上海的茶多销往国外。这个选项没有在背景信息中得到支持，背景只提到了上海是茶叶的消费与流通中心，并未提及茶叶主要销往国外。
B. 上海闸北区种植茶叶。这个选项与背景信息相矛盾，背景明确提到上海不产茶叶，因此闸北区也不可能种植茶叶。
C. 上海的茶叶消费量很大。这个选项与背景信息一致，背景明确指出上海是中国最大的茶叶消费城市，说明其茶叶消费量很大。
D. 上海是中国最大的茶叶产地。这个选项与背景信息相矛盾，背景明确提到上海不产茶叶。\newline
3.综合以上分析，我们可以得出结论，与试题内容一致的一项是C. 上海的茶叶消费量很大。
因此得出，答案：C. 上海的茶叶消费量很大。
\\ \hline
72B & \footnotesize 根据背景信息，上海不产茶叶，但是它是中国最大的茶叶消费与流通城市。这表明上海内部对茶叶有着巨大的需求，即消费量很大。选项A提到了销往国外，但背景中没有提及；选项B提到了种植茶叶，与背景中上海不产茶叶的事实矛盾；选项D则错误地声称上海是中国最大的茶叶产地，与背景信息相反。因此，根据上海是中国茶叶消费和流通中心的事实，可以推断出正确答案是C，即上海的茶叶消费量很大。
因此得出，答案：C. 上海的茶叶消费量很大。 \\ \hline
7B-Seen & \footnotesize 首先，我们需要仔细分析题目中给出的背景信息。背景提到，上海不产茶叶，但却是中国最大的茶叶消费与流通城市。这说明上海的茶叶消费量非常大，远超其他地区。\newline
接下来，我们逐一考虑各个选项：
A. 上海的茶多销往国外。这个选项并没有在背景信息中得到明确的支持，所以不能确定其正确性。
B. 上海闸北区种植茶Ⅱ十。这个选项与背景信息相矛盾，因为背景明确提到上海不产茶叶，所以这个选项是错误的。
C. 上海的茶叶消费量很大。这个选项与背景信息中的“上海不产茶叶，却是中国最大的茶叶消费与流通城市”相吻合，因此这个选项是正确的。
D. 上海是中国最大的茶叶产地。这个选项与背景信息相矛盾，因为背景明确提到上海不产茶叶，所以这个选项是错误的。\newline
综合以上分析，我们可以得出结论：上海的茶叶消费量很大。因此得出，答案：C. 上海的茶叶消费量很大。 
 \\ \hline
7B-Unseen & \footnotesize 1.首先，我们需要理解问题的背景。背景中提到，上海不产茶叶，却是中国最大的茶叶消费与流通城市。这说明上海的茶叶消费量很大。\newline
2.接下来，我们分析选项：
A. 上海的茶多销往国外：这个选项没有提到茶叶的消费量，所以不能确定。
B. 上海闸北区种植茶Ⅱ十：这个选项与问题背景无关，所以不能确定。
C. 上海的茶叶消费量很大：这个选项与问题背景相符，说明上海的茶叶消费量很大。
D. 上海是中国最大的茶叶产地：这个选项与问题背景相矛盾，因为背景中提到上海不产茶叶。\newline
综合以上分析，我们可以得出结论：与问题内容一致的答案是C. 上海的茶叶消费量很大。因此得出，答案：C. 上海的茶叶消费量很大。
\\ \hline
\end{tabular}
\caption{Original Chinese cases in Table \ref{app:rationale_comp}}
\label{app:rationale_comp_zh}
\end{table*}

\end{CJK}

\end{document}